\documentclass[lettersize,journal]{IEEEtran}
\usepackage{amsmath,amsfonts}
\usepackage{algorithmic}
\usepackage{algorithm}
\usepackage{array}
\usepackage{subfig}
\usepackage{textcomp}
\usepackage{stfloats}
\usepackage{url}
\usepackage{verbatim}
\usepackage{graphicx}
\usepackage{subcaption}
\usepackage{enumitem}
\usepackage{cite}
\usepackage{color}
\usepackage{multirow}
\usepackage{booktabs}
\usepackage{makecell}
\usepackage{xcolor}
\usepackage{hyperref}
\usepackage{caption}
\usepackage{siunitx}  % Required for alignment
\begin{document}

\title{Subjective Portrait Region Cropping in Landscape Videos with Temporal Annotation Smoothing}

%\author{Bowen Chen, Cheng-han Lee, Yixu Chen, Zaixi Shang, Hai Wei, and Alan C. Bovik, ~\IEEEmembership{Life Fellow,~IEEE}}
      
\author{Cheng-Han Lee, Maniratnam Mandal, Neil Birkbeck, Yilin Wang, Balu Adsumilli, ~\IEEEmembership{Fellow,~IEEE}, Alan C. Bovik, ~\IEEEmembership{Life Fellow,~IEEE}} %\thanks{}     
        
        % <-this % stops a space
% \thanks{This paper was produced by the IEEE Publication Technology Group. They are in Piscataway, NJ.}% <-this % stops a space
% \thanks{Manuscript received April 19, 2021; revised August 16, 2021.}}

% The paper headers
\markboth{Journal of \LaTeX\ Class Files,~Vol.~XX, No.~X, MONTH~20XX}%
{Shell \MakeLowercase{\textit{et al.}}: A Sample Article Using IEEEtran.cls or IEEE Journals}

%\IEEEpubid{0000--0000/00\$00.00~\copyright~2021 IEEE}
% Remember, if you use this you must call \IEEEpubidadjcol in the second
% column for its text to clear the IEEEpubid mark.

\maketitle

\begin{abstract}
With the rise of mobile video consumption on diverse handheld display resolutions and orientation modes, altering videos to aspect ratios poses challenges. Static cropping and border padding often compromises visual quality, while warping may distort a video's intended meaning. Here we advocate for a more effective approach – cropping significant regions within video frames in a temporal manner, while minimizing distortion and preserving essential content. One barrier to solving this problem is the lack of sufficiently large-scale database devoted to informing these tasks. Towards filling this gap, we introduce the LIVE-YouTube Video Cropping (LIVE-YT VC) database, featuring 1800 videos, annotated by 90 human subjects. Using videos sourced from the YouTube-UGC and LSVQ Databases, this new resource is the largest publicly-available subjective video portrait region cropping database. 
We also introduce a post-processed version of the database, called LIVE-YT VC++, whereby a novel intra-frame temporal filter was deployed to smooth subjective annotations within each video.
We demonstrate the usefulness of this new data resource using the SmartVidCrop \cite{apostolidis2021fast} algorithm and state-of-the-art video grounding models, in hopes of establishing our subjective dataset as a benchmark for future research. Our contributions offer a resource for advancing video aspect ratio transformation models towards ensuring that reshaped mobile-friendly video content retains its quality and meaning. Since our labels bear resemblances to video saliency annotations, we also conducted an additional analysis to explore the similarity between our labels and video saliency predictions. Finally, we repurposed state-of-the-art video grounding models for aspect ratio change tasks, and fine-tuned them on our dataset. As a service to the research community, we plan to publicly release the videos, the metadata of the new database, source code, and video demo at  \href{https://github.com/steven413d/LIVE-YT-VideoCropping}{\textcolor{blue}{https://github.com/steven413d/LIVE-YT-VideoCropping}}.
\end{abstract}


\section{Introduction}
\IEEEPARstart{T}{he} increasing use of mobile devices for video consumption has led to the need for videos in specific aspect ratios that differ from the traditional 16:9 used for home television viewing. To comply with video-sharing platforms’ requirements, existing videos often need or would benefit by being transformed to different aspect ratios. However, simple approaches such as static cropping or padding with black borders often lead to a significant losses of visual content or unsatisfactory experiences. Common methods like warping can introduce distortions or alter a video’s meaning. In our view, a better approach is findind and dynamically cropping significant regions in video frames is more suitable for video aspect ratio transformation. Properly done, this can minimize spatial distortions, while capturing the visually important parts of video frames as they play. Since this problem is complex, relating as it does to visual saliency and requirements of temporal continuity, data-driven approaches are likely necessary. Unfortunately, there is currently a lack of any large database of human-selected crops of original video clips. The only somewhat relevant existing human-annotated database that we could find, called RetargetVid \cite{apostolidis2021fast}, contains only 200 videos. Also, its annotations only support crops that preserve the original height.
Another new database, called LOPOV \cite{gu2025dynamic}, leveraged a vision-language model and an image composition model to auto-generate its dataset. However, it ignores human preference because the data is labeled solely by the model.
The limited data in RetargetVid makes it difficult to use it design deep learning-based video cropping models. Towards filling this gap, we created a new database called the LIVE-YouTube Video Cropping (LIVE-YT VC) Database. LIVE-YT VC contains 1800 videos, each cropped by 30 subjects. 

However, since each video in the LIVE-YT VC database was labeled by 30 different human subjects, there are naturally temporal disagreements among them, as different individuals may focus on and select different regions of interest (ROI) at each moment within a same video. Furthermore, to ensure reasonable allocations of human effort, each frame was annotated by a single subject, resulting in sparse labeling crops. To address this issue, we also developed a novel weighting-based bilateral temporal filtering mechanism which ensures smooth cropping flows, while effectively aggregating the assigned crops. As a result, each frame reflects a combined input from approximately 8.12 subjects.

%-----------------------------------------------------------%
\begin{figure}[t]
\begin{center}
 \includegraphics[width=1.0\linewidth]{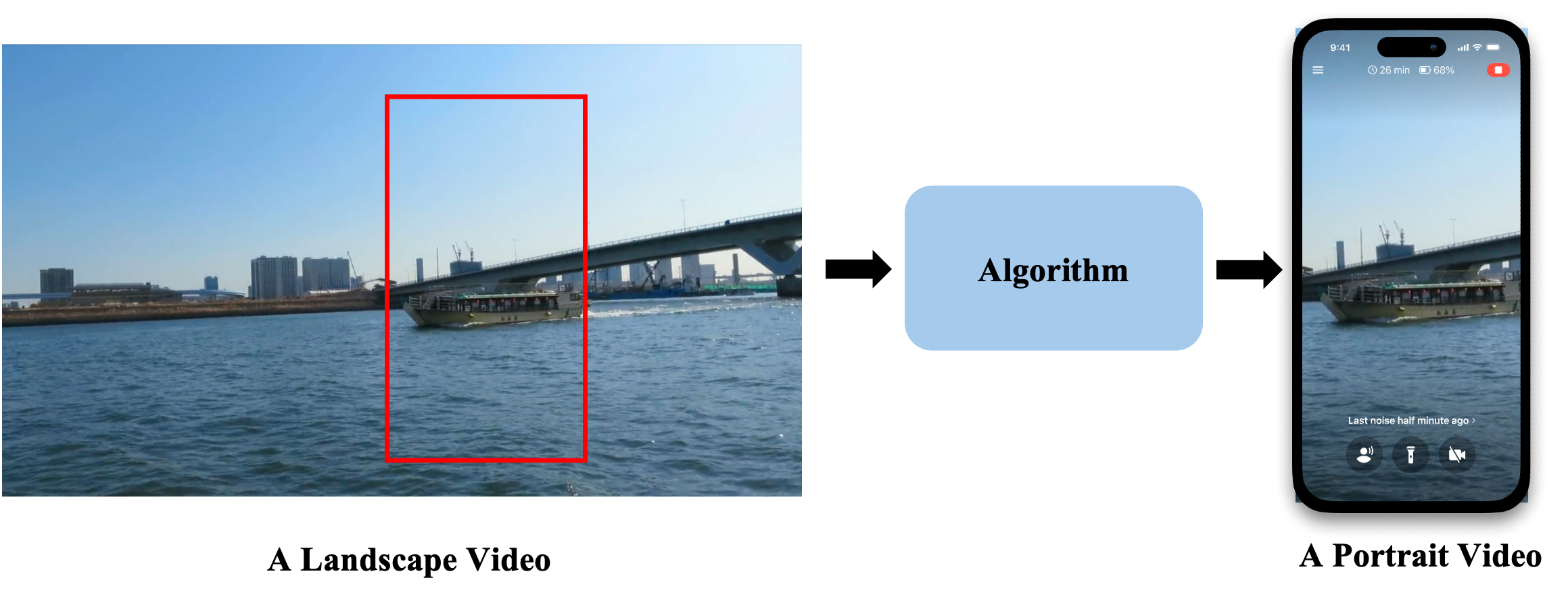}
\end{center}
\vspace{-10pt}
\caption{\footnotesize Landscape video aspect ratio transformation system. Given one lanscape video, the algorithm can automatically transform the video into a portrait like video tube.}
\label{fig:long}
\label{fig:onecol}
\label{fig1}
\vspace{-5pt}
\end{figure}  
%-----------------------------------------------------------%

The contributions we make are six-fold:
\begin{itemize}
\item We conducted the largest subjective video portrait region cropping study to date,  involving 90 subjects who deliver 54,000 portrait crops of landscape videos (Sec. \ref{sec:human_study}).
\item The outcome of the human study is the largest subjective video portrait region cropping database, consisting of 1800 videos sampled from the YouTube-UGC \cite{wang2019youtube} and the LIVE-FB Large-Scale Video Quality (LSVQ) \cite{ying2021patch} databases. This new perceptual data resource contains 324,000 frames extracted from the videos. We also conduct a holistic data analysis on the subjective data (Sec. \ref{sec:data_analysis}).
\item We introduce a novel intra-frame temporal filter that smoothes subjective annotations collected on each video in the LIVE-YT VC database, yielding an enhanced version that we call LIVE-YT VC++.  (Sec. \ref{sec:liveytvcplus}). Both the original and enhanced versions are being made publicly available.
\item We conducted an analysis of the relationship between visual saliency and the cropping labels supplied by the human participants, on both LIVE-YT VC and LIVE-YT VC++ (Sec. \ref{sec:experiments}).
\item We studied the usefulness of the new databases by testing the saliency-based algorithm SmartVidCrop \cite{apostolidis2021fast} on them. (Sec. \ref{sec:experiments})
\item We fine-tuned two state-of-the-art video grounding models on the new databases, to evaluate their accuracy and visual quality. (Sec. \ref{sec:experiments})
\end{itemize}  
    
\section{Related Works}
We briefly review two pertinent research topics: image cropping and video retargeting. Unlike traditional image cropping, our focus lies in addressing both spatial and temporal challenges that arise when processing videos. Unlike conventional video retargeting, our approach is specialized to computing videos having the fixed aspect ratio 9:16, which aligns with the prevalent aspect ratio used by contemporary handheld devices.
\subsection{Image Cropping}
Image cropping aims to improve the composition as well as the aesthetic quality of an image by removing extraneous content. It is widely used in photography, printing, and thumbnail generation. Notable algorithms within the domain of general image cropping include \cite{li2018a2, wang2017deep, zeng2019reliable}. For aspect-ratio selectable image cropping, relevant contributions include \cite{esmaeili2017fast, kishore2019user, li2020learning}. % image based method, need to compare? % more detail about each method
\subsection{Video Retargeting}
The realm of video retargeting encompasses distinct categories such as warping, cropping, and seam carving. In the following, we delve into key works within each of these categories.
\subsubsection{Cropping Based methods}
Cropping-based techniques rely on identifying an area within each analyzed video frame deemed to have the greatest visual significance, then cropping frames accordingly. This proves effective when each frame contains a single primary object. However, challenges arise on frames containing multiple important objects, risking the inadvertent cropping of some of them. A notable method in this category is SmartVidCrop \cite{apostolidis2021fast}, which introduced the first video cropping benchmark. SmartVidCrop employs a saliency map alongside a novel clustering algorithm to identify a significant region. \cite{gu2025dynamic} introduces LOPOV , a large-scale, composition-aware dataset auto-generated using vision-language and image composition assessment models. Based on this data, they propose the FEVR end-to-end framework. However, the authors do not open-source their model or code.
\subsubsection{Warping Based Methods}
Warping-based retargeting focuses on first identifying important regions (ROIs), then subsequently shrinking unimportant areas. Unlike cropping methods, warping handles multiple important regions without introducing distortion. However, artifacts may appear in the shrunken unimportant regions, and significant computational effort is required to reliably conduct ROI estimation. Noteworthy works in this category include \cite{wolf2007non} and \cite{krahenbuhl2009system}.
\subsubsection{Seam Carving Methods}
Seam carving techniques involve the removal of paths consisting of pixels having low importances, as determined by low gradient values. This approach is suitable for images such as landscapes. Seam carving may result in distortion or artifacts due to the appearance of new edges, and it is not ideal for images containing many large gradients. The works reported in \cite{rubinstein2008improved} contributed significantly to this approach.
\subsubsection{Hybrid Methods}
Hybrid methods merge the benefits of cropping, warping, and seam carving to find a most suitable combination for retargeting. However, this method requires also considerable computation. One notable work in this area is \cite{rubinstein2009multi}.

In summary, from among the various approaches, cropping-based methods are more suitable for conducting video aspect ratio transformation, as it can potentially minimize distortions, while preserving essential content. Of course, this would require correctly and consistently identifying the important content, not only over space, but smoothly over time.
\section{Details of the Subjective Study}
\label{sec:human_study}
%-----------------------------------------------------------%
\begin{figure*}[t]
\begin{center}
 \includegraphics[width=1.0\linewidth]{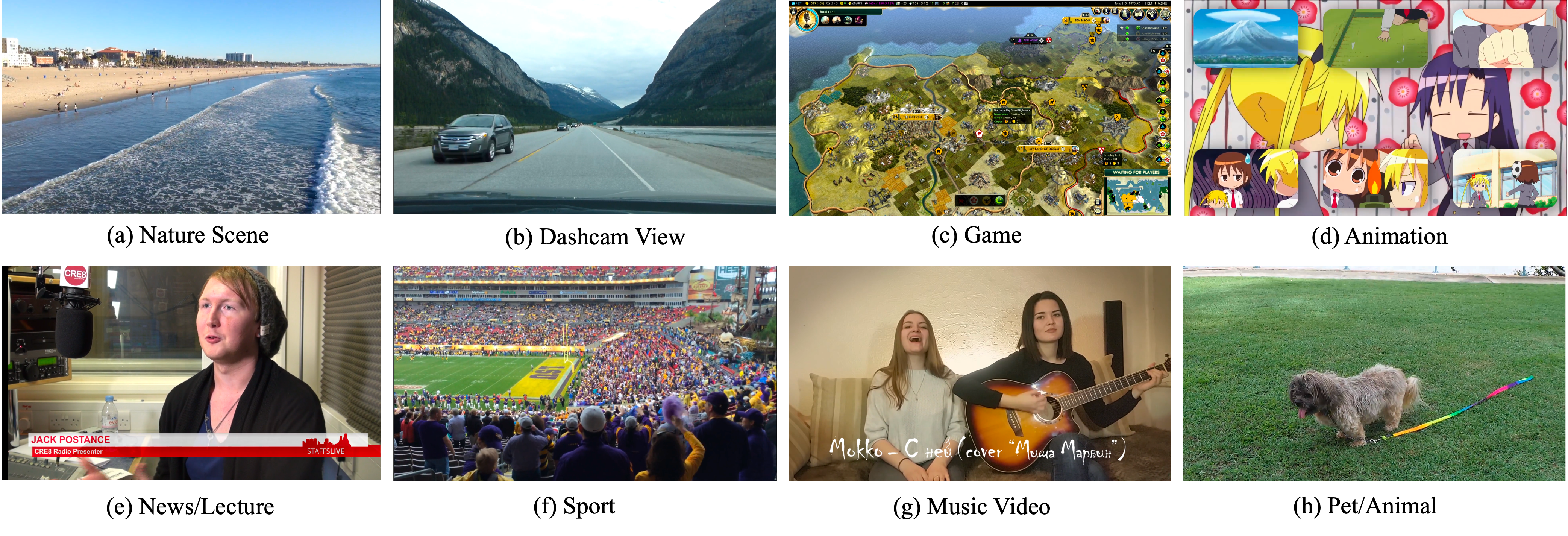}
\end{center}
\vspace{-10pt}
\caption{\footnotesize Sample frames of landscape videos drawn from the LIVE-YT VC Database.}
\label{fig:long}
\label{fig:onecol}
\label{fig2}
\vspace{-5pt}
\end{figure*}  
%-----------------------------------------------------------%
\begin{figure}[t]
\begin{center}
 \includegraphics[width=1.0\linewidth]{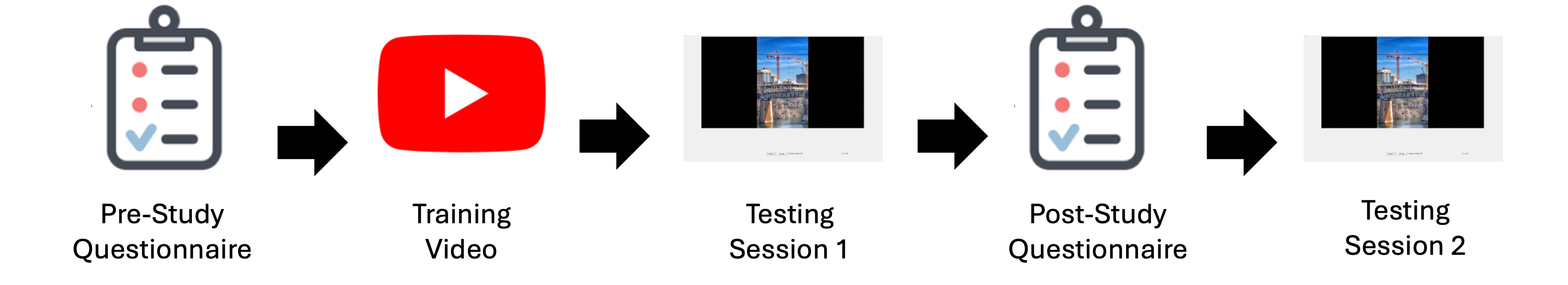}
\end{center}
\vspace{-10pt}
\caption{\footnotesize This diagram illustrates the human study workflow.}
\label{fig:long}
\label{fig:onecol}
\label{fig3}
\vspace{-5pt}
\end{figure}  
%-----------------------------------------------------------%
Next we provide a comprehensive overview of the human study we conducted, including all procedural details and data labeling. The human annotations are in the form of bounding boxes, indicating those frame regions that the participants found to be the most interesting, important, or good aesthetic composition.
\subsection{Data Source}
\label{ssec:subhead}
Our video data was sampled from two video quality assessment benchmarks -- the YouTube-UGC \cite{wang2019youtube} Database and the Large-Scale Social Video Quality (LSVQ) \cite{ying2021patch} database. Sample frames from some of the selected source videos are shown in Fig. \ref{fig2}.
\subsubsection{YouTube-UGC Database}
\label{sssec:subsubhead}
The YouTube-UGC dataset is a publicly accessible UGC-VQA database. It comprises 1,380 20-second video clips sampled from a diverse array of YouTube videos, numbering in the millions. The videos in the dataset span various resolutions, including 360p, 480p, 720p, 1080p, and 4k. YouTube-UGC is also systematically organized into 15 distinct categories, covering Animation, Cover Songs, Gaming, HDR, How To, Lectures, Live Music, Lyric Videos, Music Videos, News Clips, Sports, Television Clips, Vertical Videos, Vlogs, and VR. The videos in this dataset were quality rated by more than 8,000 human subjects.

\subsubsection{Large-Scale Social Video Quality (LSVQ) database}
\label{sssec:subsubhead}
The LIVE-FB Large Scale Video Quality (LSVQ) dataset is composed of 39,000 real-world distorted videos sourced from the Internet Archive and YFCC-100M \cite{thomee2016yfcc100m}. This extensive dataset also incorporates 5.5 million human perceptual quality annotations. The video durations within LSVQ span range from 5 to 12 seconds, covering a range of video resolutions: 240p, 360p, 480p, 720p, and 1080p.
\subsection{Video Sampling and Pre-processing}
\label{ssec:subhead}
We limited the selection of videos to those having durations exceeding 6 seconds (or 180 frames) and having a resolution of 1080p. From the YouTube-UGC dataset, we ended up with 224 videos meeting these criteria. From the LSVQ dataset, we randomly sampled 1576 videos from a pool of 2,582 available videos meeting the aforementioned criteria. Each of these 1800 selected videos was randomly cropped to temporal duration of 6 seconds.

\subsection{Subject Screening and Training}
\label{sec:trainging}
We conducted a human study whereby 90 volunteer human subjects labeled the video frames. Each participant was asked to undergo five tasks, as depicted in Fig. \ref{fig1}. In the pre-study stage, participants were asked to complete a questionnaire that explained the study objectives, ensured they have normal or corrected-to-normal vision, and inquired about their willingness to provide their data for research purposes.
Next, during a subject training stage, each participant was presented with a video designed to familiarize them with the study's objectives and guide them on how each button functions and how to use the mouse to select bounding boxes. Following this, each participant engaged in two separate sessions, on separate days. At the end of the first session, each participant was prompted to complete a post-study questionnaire.

%-----------------------------------------------------------%
\begin{figure}[t]
\begin{center}
 \includegraphics[width=1.0\linewidth]{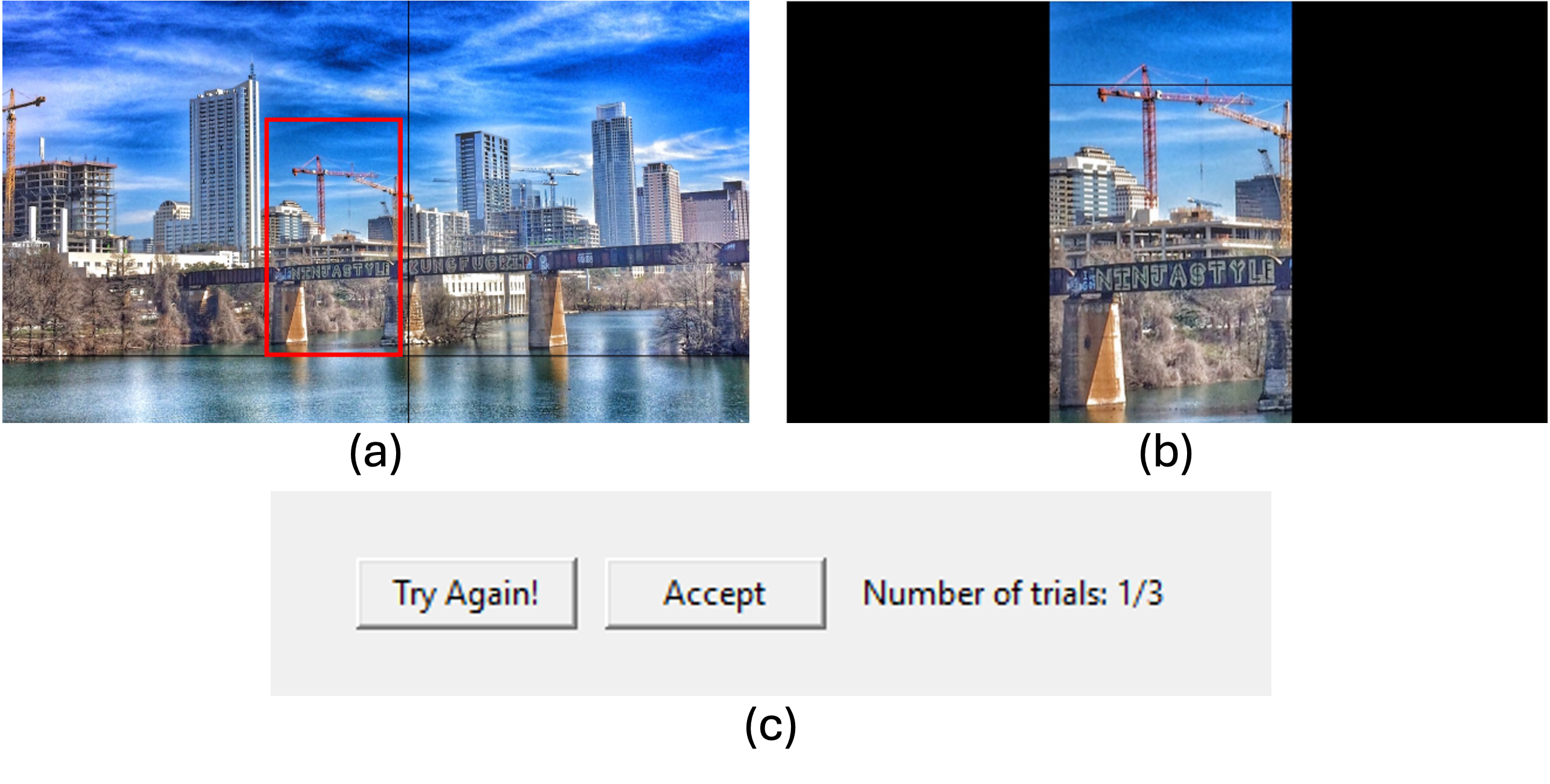}
\end{center}
\vspace{-20pt}
\caption{\footnotesize Visualization of the visual cropping task each human subject participated in. (a) shows the labeling mode of the interface. (b) illustrates the visualization mode when the subject clicks `Accept'. The cropped region is resized to meet the largest vertical dimension. (c) displays the two interactive buttons available for users: 'Try Again!' and 'Accept'.}
\label{fig:long}
\label{fig:onecol}
\label{fig2}
\vspace{-10pt}
\end{figure}
%-----------------------------------------------------------%

\subsection{Study Protocol}
\label{ssec:subhead}
The study consisted of two sessions, where each session consisting of 300 presented video frames (all from different videos), totaling 600 video frames being rated by each subject. The participant pool comprised 90 individuals, organized into three groups, with each group containing 30 subjects. Each session was of duration of about 45 minutes. A total of 30 frames were sampled from each video every 6 frames, and each frame was rated by a single rater, yielding a total of 30 ratings per video.

\subsection{Study Environment}
\label{ssec:subhead}
The study was conducted in The University of Texas at Austin, LIVE human study room. Two 27-inch 4K monitors were used to present the videos for subject annotation. Each monitor was connected to a workstation having the Windows 10 operating system. Each user was provided with a mouse and keyboard which connected to the workstation to operate the interface.

\subsection{Study Interface}
\label{sec:interface}
Fig. \ref{fig2} shows the two modes of the interface - labeling mode and visualization mode. In the labeling mode, the interface displays a single video frame. After the user interactively drags a bounding box on the image to crop a desired region, the resulting crop was resized, without changing the aspect ration of the crop using Lanczos interpolation \cite{lanczos} to match the vertical dimension of the screen. The user was then given two options: 
\textbf{(1) Try Again!} If the cropped and automatically resized portrait mode image did not satisfy them or was not of good enough quality, the subject could click the button to try again. Subjects were limited to three tries per image. If the user tried three times, the system accepted the third attempt and automatically proceeded to the next image. 
\textbf{(2) Accept.} If satisfied with their current upsampled portrait mode image (the user did not have to try 3 times), clicking `Accept' would let the subject move forward to the next image.

\subsection{Post Study Questionnaire}
\label{sec:post_study}
At the end of each rating session, participants completed a questionnaire to provide demographic information, feedback on the study protocol, and comments regarding the ease of participation. Approximately 74\% of the participants were male, and 26\% were female. Participant ages ranged from 19 to 31 years, with a mean of 22.91, a median of 21, and a standard deviation of 7.99. About 98\% of the participants were students at The University of Texas at Austin. None reported experiencing dizziness during the sessions. All participants had normal or corrected-to-normal vision, and none were color blind.

\subsection{Comparison Against Existing Human Annotated Database}
\label{sec:compare_database}
Table.\ref{tab1} compares the existing RetargetVid \cite{apostolidis2021fast} against the new LIVE-YT VC dataset.  RetargetVid employed a smaller panel of six raters, with dense crops provided on every frame. Our database employed videos of a higher resolution of 1080p than RetargetVid. Whereas RetargetVid offers two choices of cropped aspect ratios, the human participants in our study were required to supply the prevalent device portrait aspect ratio 9:16. Lastly, unlike RetargetVid, our database did not constrain the crops to preserve frame heights, affording the subjects greater flexibility in selecting their visually important regions.
%-----------------------------------------------------------%
\begin{table}
\caption{Comparison between LIVE-YT VC and existing human annotated database.}
\footnotesize
\begin{center}
\begin{tabular}{|c|c|c|} 
\hline
 & RetargetVid \cite{apostolidis2021fast} & LIVE-YT VC \\ 
\hline
\hline
\# Videos & 200 & 1800\\
\hline
Video length & 17 to 42 seconds & 6 seconds \\
\hline
\# Subjects per video & 6 & 30 \\
\hline
\# Ratings per video & 6120 to 15120 & 30 \\
\hline
Skip frames & 0 frames & 6 frames\\
\hline
Resolution & 360p & 1080p\\
\hline
Frame rates (fps) & 30 & 30\\
\hline
Data sources & DHF1K \cite{wang2018revisiting} & YouTube-UGC \cite{wang2019youtube}, LSVQ \cite{ying2021patch}\\
\hline
Aspect ratios & 1:3, 3:1 & 9:16\\
\hline
Flexible resolution & No & Yes\\
\hline
\end{tabular}
\end{center}
\label{tab1}
\vspace{-2em}
\end{table}
%-----------------------------------------------------------%
\section{Data analysis}
\label{sec:data_analysis}
We collected 54,000 bounding box labels, one per frame, hence 30 per video. We conducted a statistical analysis on the outcomes. The following sections outline our data rejection methodologies and analyses of subjective annotations.

\subsection{Content Diversity}
\begin{figure}[t]
\begin{center}
 \includegraphics[width=1.0\linewidth]{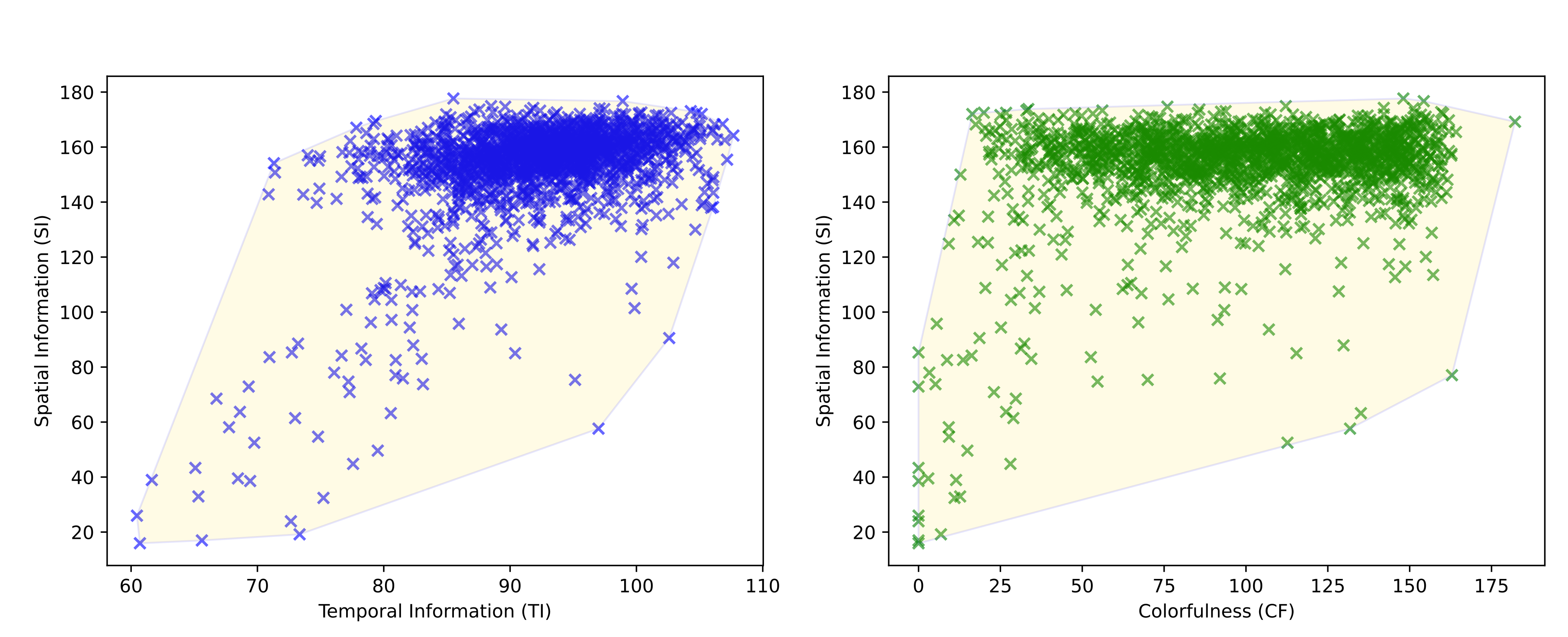}
\end{center}
\vspace{-10pt}
\caption{\footnotesize Spatial Information (SI) plotted against Temporal Information (TI) (Left) and Spatial Information (SI) plotted against Colorfulness (CF) on the source videos (Right).}
\label{fig:diversity}
\vspace{-5pt}
\end{figure}

To capture the diversity of video content in our database, we computed three objective features on all 1,800 videos: Colorfulness \cite{hasler2003measuring}, Spatial Information (SI), and Temporal Information (TI), as recommended in \cite{winkler2012analysis}, \cite{itu_p910}. Colorfulness and SI were calculated on a per-frame basis, then averaged across frames, while TI was derived from frame differences. The final feature values are shown in Fig. \ref{fig:diversity}, illustrating a wide range of spatial and temporal complexities, as well as variations in colorfulness, illustrating the diversity of the dataset.

%-----------------------------------------------------------%
%-----------------------------------------------------------%

\subsection{Data Rejection}
Data rejection can be studied from two perspectives -- subject rejection due to dishonest ratings, and outlier analysis. 
\subsubsection{Subject Rejection}
\begin{figure}[t]
\begin{center}
 \includegraphics[width=1.0\linewidth]{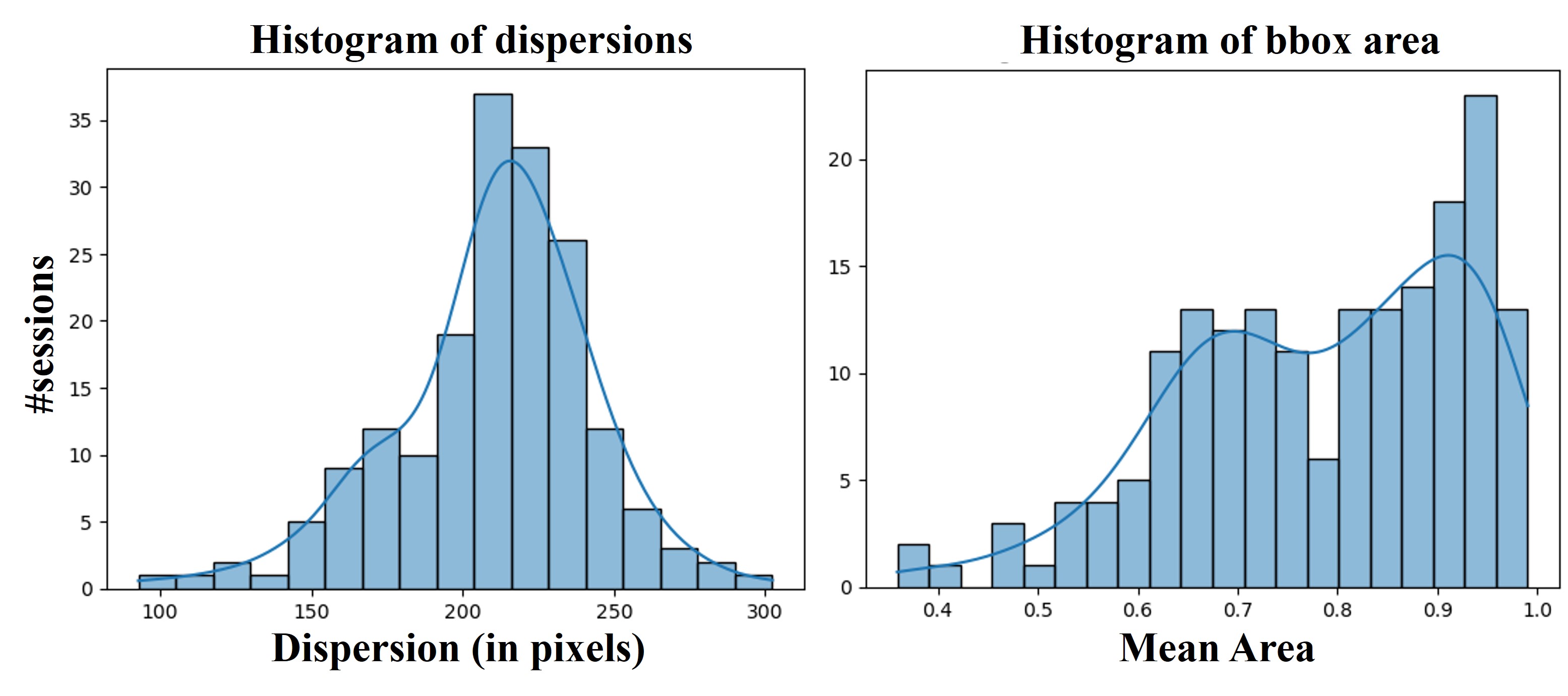}
\end{center}
\vspace{-10pt}
\caption{\footnotesize Left: plot of the histogram of dispersion values over all sessions. Right: plot of the histogram of the mean bounding box area (rescaled) per session over all sessions.}
\label{fig:sub_rej}
\vspace{-5pt}
\end{figure}

When participating in their study sessions, each subject was required to label 300 video frames. We analyzed the bounding box labels to determine subjects whose scores should be rejected as outliers or did not participate honestly. To rule out similar ratings, that is a subject labeling the same region on all frames, we calculated the dispersion of bounding box centers and plotted the histogram over all 180 sessions. We hypothesized that if a subject labeled all the frames similarly, the dispersion value of box centers would be very small within a session. The distribution of dispersion values for each session is shown in Fig. \ref{fig:sub_rej} (left). We found that the least dispersion value was around 100 pixels which we deemed to be reasonable for bounding box labels on 1080p frames. 

%-----------------------------------------------------------%
%-----------------------------------------------------------%

We also checked whether a subject labeled only small bounding boxes on all frames in their session (which would mean a poor understanding of the study objective) by calculating the mean bounding box area over a session, and plotting the histogram over all 180 sessions. It may be observed from Fig. \ref{fig:sub_rej} (right) that most sessions produced a high average box area, with minimum area around 0.4 (normalized). Hence we concluded that none of the subjects gave bad labels voluntarily.

\subsubsection{Outlier Analysis}
Because we obtained only one rating per frame, we carried out an outlier analysis on the 30 ratings per video and considered the bounding box centers to be the sample data points. To justify the validity of such an analyses the frames sampled from a video would be assumed to be similar enough to garner similar labels. Hence we excluded all videos containing very high motion or multiple scenes, ending up with a subset of 1,566 videos. We used the Pyscenedetect library to determine the number of scene changes in each video content. The two methods of outlier detection we pursued were the Local Outlier Factor (LOF) method \cite{lof} and the Z-score method \cite{modzscore}. 

The LOF (Local Outlier Factor) utilizes a nearest neighbors approach to detect outliers, by assigning each data point a score indicating its isolation within a local neighborhood. A higher score indicates a larger neighborhood size and a greater likelihood of being an outlier. In the Z-score method, we normalized the bounding box centers, then rejected points based on their estimated distributions. We detected 3,035 (6.5\%) and 2,724 (5.8\%) outlier labels out of 47K ratings using the LOF and Z-score methods, respectively. 

\subsection{Inter-subject Labeling Consistency}
\label{ssec:inter-consistency}
Being constrained by one label per frame, we had to develop other methods to test inter-subject consistency than are popularly used in subjective quality studies. 

\subsubsection{Intersection over Union (IoU) Analysis}
\begin{figure}[t]
\begin{center}
 \includegraphics[width=1.0\linewidth]{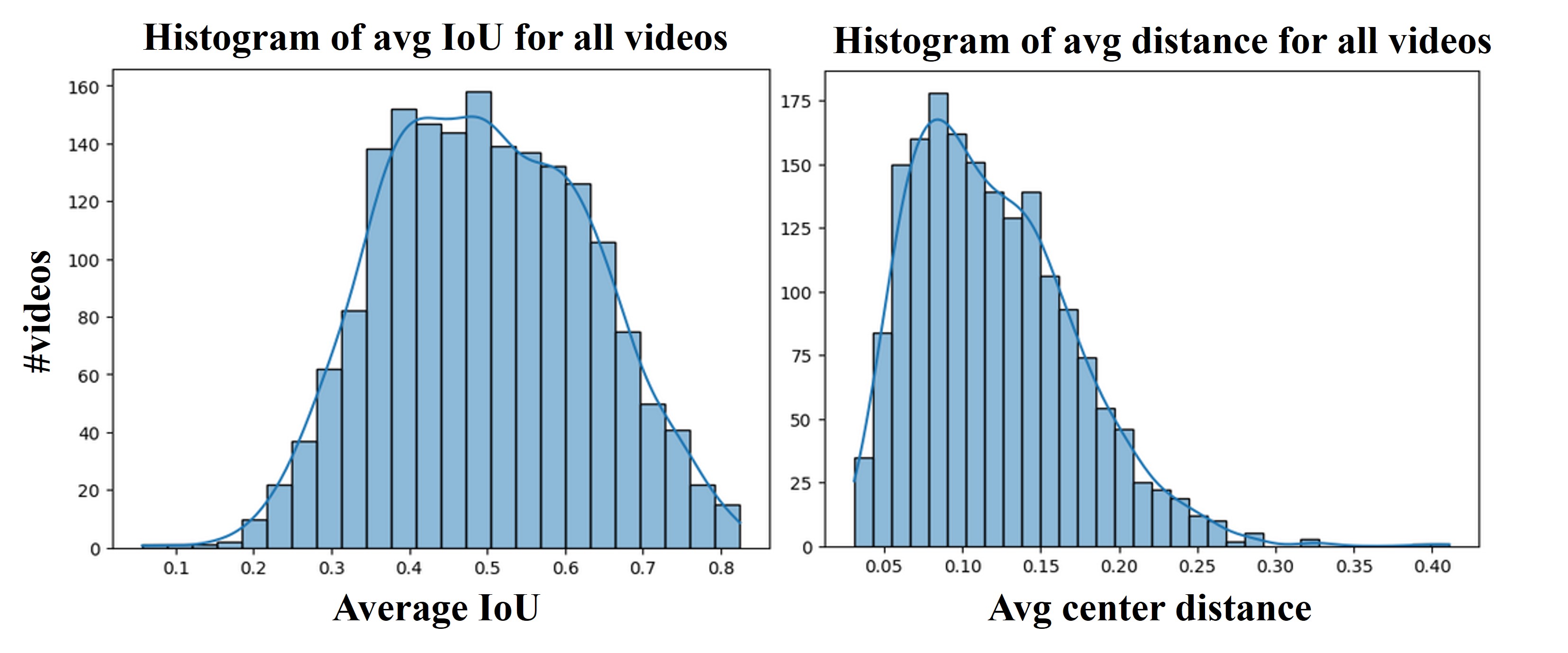}
\end{center}
\vspace{-10pt}
\caption{\footnotesize We observe a skew towards higher values of average IoU across the dataset (left) and a skew towards lower values for the distribution of bounding box center (rescaled) distances (right), indicating high subject labeling consistency across the dataset.}
\label{fig:bbox_const}
\vspace{-5pt}
\end{figure}

Excluding high-motion videos or videos with multiple scene changes (13\% of our dataset), we expect a video to have similar subjective labels across its temporal dimension. With that assumption, consecutive frames should have high IoU, i.e. high overlap of the labeled bounding boxes. Given that each frame of a video was rated by different subjects, we calculated the IoU on consecutive frame pairs and computed the average IoU across each video. From Fig. \ref{fig:bbox_const} (left), it may be observed that most videos have significant IoU between frame labels, and the average IoU across all videos was $0.50$, which signifies an overall good consistency of the subject labels.  

%-----------------------------------------------------------%
%-----------------------------------------------------------%

\subsubsection{Center Distance Analysis}
Similar to the IoU analysis, which measured the overlap of bounding boxes for consecutive frame pairs, we hypothesized that such pairs should also have little distance between the box centers. A lower average distance between the centers across a video should indicate a higher consistency among subject labels. Fig. \ref{fig:bbox_const} (right) shows the histogram of the box center distances across all videos in the dataset. We noticed that the distribution is skewed toward lower values, indicating a higher consistency of subject labels.      

\subsection{Bounding Box Statistics}
\label{ssec:bbox_stat}
To gather intuition on how the bounding boxes were selected by the subjects, we computed other general statistics on all the videos in the dataset.

\subsubsection{Center Distance}
\begin{figure}[t]
\begin{center}
 \includegraphics[width=1.0\linewidth]{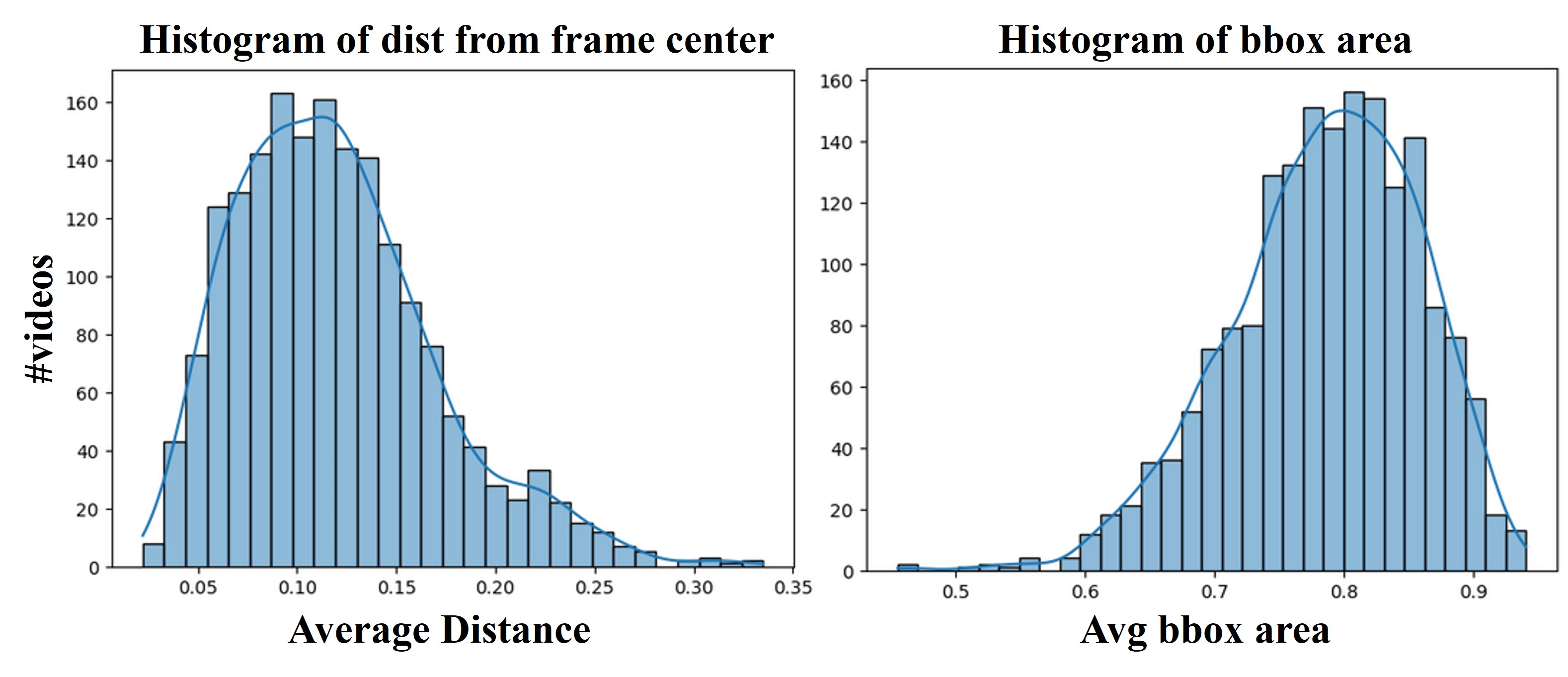}
\end{center}
\vspace{-10pt}
\caption{\footnotesize We observe a skew towards lower values of the average distance (rescaled) of bounding box centers from frame centers (left) and a skew towards higher values for the distribution of bounding box areas (rescaled) distances (right), indicating labeling trends across the dataset.}
\label{fig:data_stats}
\vspace{-5pt}
\end{figure}

On each frame, we computed the distance of the labeled bounding box center from the frame center and average it over all frames in each video. The histogram plot of average center distances shown in Fig. \ref{fig:data_stats} (left) reveals that most videos were skewed toward lower distance values, indicating that the subjects had a high inclination to choose the central region of a video when cropping portraits (center bias).  

%-----------------------------------------------------------%
%-----------------------------------------------------------%

\subsubsection{Box Area}
We also computed the areas of the bounding boxes in each frame and averaged them across each video. The histogram of the average area is shown in Fig. \ref{fig:data_stats} (right). We found that most videos were labelled with large bounding boxes, implying that the subjects tended to select large regions, perhaps attempting to capture more information, when cropping the portrait shaped images.

\section{LIVE-YT VC++}\label{sec:liveytvcplus}
\subsection{Intra-frame Temporal Filter Algorithm}
%To conduct the intra-frame temporal filtering, firstly, we need to define how to represent a box annotation.cd 
% Def of the box. Introduce box representation (x, y, r)
% (1) Improve the label consistency.
% (2) Improve the subjive number in each frame.
% (3) Reduce the noise due to subject bias.
% weight sampled functions -> hamming window : visualize
% Assumption -> frame distance.

% Hanming window formula here!

Although the LIVE-YT VC database includes labels from 30 subjects per video, each frame is annotated by only a single person, leading to sparse and frequently temporally inconsistent labels due to varying attention and/or regions of interest. To address this, we have devised a novel weighting-based bilateral temporal filtering method that effectively smooths annotations and aggregates opinion.

\subsubsection{Box Representation} \label{sec:box_represent}
Before introducing our filtering model, we first define box series annotations within a video.
\begin{equation}
\begin{split}
Box^{j}_{i} = (x^{j}_{i},\: y^{j}_{i},\: r^{j}_{i}),\:where\:i \in [1, 30]\: and\: j \in [1, 1800]
\label{eq1},
\end{split}
\end{equation}
In this equation, $i$ denotes the box ID within a video, while $j$ represents the video ID within the dataset. $x$ and $y$ refer to the center coordinates of the annotation, while $r$ is the size ratio - defined as the annotation height divided by 1080. Note that all of the videos in our dataset are 1080p.

%\begin{algorithm}[H]
%\footnotesize
%  \caption{Intra-frame Annotation Temporal Filtering}
%  \label{alg1}
%  {\bf Input:}
%    $I^{t}, M^{t}, M^{ref}$
    
%  {\bf Output:}
%    $I^{out}, I^{blend}$
%  \begin{algorithmic}[1]
%  \WHILE{iteration not converge}
%  \STATE Choose one minibatch of $N$ mask and image pairs $\left \{M^{t}_i, M^{ref}_i, I^{t}_i \right \}$, $i=1,...,N$.
%  \STATE $z^{t} = Enc_{\mathrm{VAE}}(M^{t})$
%  \STATE $z^{ref} = Enc_{\mathrm{VAE}}(M^{ref})$
%  \STATE $z^{inter}, z^{outer} = z^{t} \pm \frac{z^{ref} - z^{t}}{\lambda_{inter}}$
%  \STATE $M^{inter} = Dec_{\mathrm{VAE}}(z^{inter})$
%  \STATE $M^{outer} = Dec_{\mathrm{VAE}}(z^{outer})$
%  \STATE Update $G_{A}(I^{t}, M^{t})$ with Eq. \ref{eq6}
%  \STATE Update $G_{B}(I^{t}, M^{t}, M^{inter}, M^{outer})$ with Eq. \ref{eq6}
%  \ENDWHILE 
%  \end{algorithmic}
%\end{algorithm}

\subsubsection{Inter Frame Temporal Filtering Algorithm} \label{sec:algo}
\begin{figure*}[t]
\begin{center}
 \includegraphics[width=1.0\linewidth]{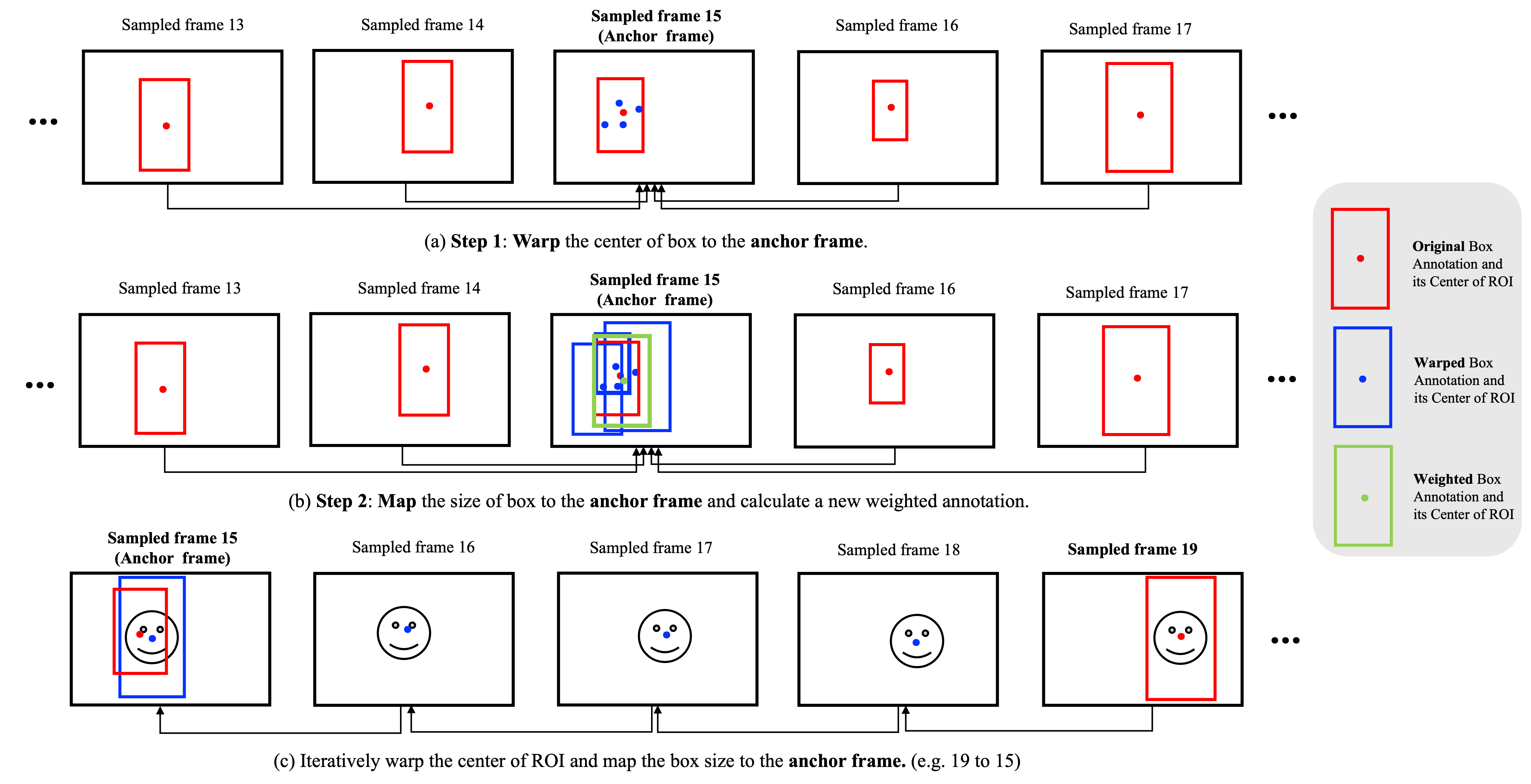}
\end{center}
\vspace{-10pt}
\caption{\footnotesize In (a), Step 1 warps the center point of the ROI annotations from adjacent frames to the anchor frame. In (b), Step 2 maps the bounding box size directly onto the anchor frame. We avoid warping the entire box to minimize distortion. In this example, the window size is 5, meaning only adjacent frames within this window are considered. In (c), we illustrate the detailed process of warping each bounding box to the anchor frame. To improve accuracy, we apply iterative warping, similar to pixel tracking techniques.} 
\label{fig:two_steps}
\vspace{-5pt}
\end{figure*}

The core idea of this algorithm is to refine annotations by leveraging temporal context. As shown in Fig. \ref{fig:two_steps} (a), the anchor frame serves as the center annotation within a filter window. This window moves across every annotation in the video. During this filtering process, each annotated frame serves as the anchor frame exactly once. The anchor frame's purpose is to aggregate the bounding boxes from its neighboring frames, which allows the system to calculate a more stable, weighted annotation based on data from multiple temporally adjacent frames. As shown in Fig. \ref{fig:two_steps} (b), given an anchor frame and its corresponding annotation, warp the annotations from adjacent frames into the anchor frame's coordinate space. Then, combine the anchor annotation with the warped annotations using a weighted averaging scheme, thereby producing a refined temporally consistent result.

\noindent\textbf{Details of ROI Center Warping and Weighting.}
To implement this, define a temporal window centered on the anchor frame, with the surrounding frames serving as context. This window slides across the 30 annotations available per video, each provided by a different subject. When creating the LIVE-YT VC++ dataset, we fited the window size to 15 frames. However, when included frames are too distant from the anchor frame, the quality of warping may degrade, especially an high-motion videos where frame content may change rapidly. To address this, we define a non-linear weighting function inspired by the Hamming window from signal processing. As the temporal distances between frames increases, the corresponding weight decreases, while the anchor frame is assigned a weight of 1. The function is defined as follows:

\begin{equation}
\begin{split}
w_{i} = 0.54 + 0.46 \times \cos (\frac{2 \times \pi \times (i - k)}{W}) 
\label{eq2}.
\end{split}
\end{equation}
In Equation \ref{eq2}, $i$ denotes the box ID of an adjacent frame, while $k$ represents the box ID of the anchor frame. $W$ refers to the size of the temporal filter window.

In Equations \ref{eq3}-\ref{eq5},  we present the formulas for computing refined annotations by combining the original box annotations with the corresponding weights. On each video and its corresponding 30 box annotations, we iterated over $k=1$ to $30$, generating a refined annotation for each frame, resulting in 30 new box annotations:

\begin{equation}
\begin{split}
x^{new}_{k} = \frac{\Sigma^{min(k+\frac{W}{2}, 30)}_{i=max(k-\frac{W}{2}, 1)} w_{i} \times x^{old}_{i}}{\Sigma^{min(k+\frac{W}{2}, 30)}_{i=max(k-\frac{W}{2}, 1)} w_{i}} 
\label{eq3},
\end{split}
\end{equation}

\begin{equation}
\begin{split}
y^{new}_{k} = \frac{\Sigma^{min(k+\frac{W}{2}, 30)}_{i=max(k-\frac{W}{2}, 1)} w_{i} \times y^{old}_{i}}{\Sigma^{min(k+\frac{W}{2}, 30)}_{i=max(k-\frac{W}{2}, 1)} w_{i}} 
\label{eq4},
\end{split}
\end{equation}

\begin{equation}
\begin{split}
r^{new}_{k} = \frac{\Sigma^{min(k+\frac{W}{2}, 30)}_{i=max(k-\frac{W}{2}, 1)} w_{i} \times r^{old}_{i}}{\Sigma^{min(k+\frac{W}{2}, 30)}_{i=max(k-\frac{W}{2}, 1)} w_{i}} 
\label{eq5}.
\end{split}
\end{equation}

Fig. \ref{fig:two_steps} (a) illustrates a visual example with a window size of 5. In the first step, the annotations from adjacent frames are warped into the anchor frame's coordinate space.

%-----------------------------------------------------------%
%-----------------------------------------------------------%

In the second step, as shown in Fig. \ref{fig:two_steps} (b), transfer the original size ratio ($r$) from the adjacent frame to the anchor frame, without warping. Only warp the center coordinates, based on the assumption that the center of the region of interest (ROI) is the primary point of focus for annotators. Warping the entire region could introduce significant distortion, especially on dynamic scenes.

Fig. \ref{fig:two_steps} (c) shows how we use the Lucas-Kanade Optical Flow algorithm \cite{baker2004lucas} to iteratively warp pixels towards the target frame, effectively tracking their motion. In our case, the tracked pixel corresponds to the center of the region of interest (ROI). We avoid directly warping pixels from the target frame to the anchor frame because, in high-motion videos, this could lead to inaccurate or failed warping due to limitations of the Lucas-Kanade algorithm.

\begin{figure}[t]
\begin{center}
 \includegraphics[width=0.9\linewidth]{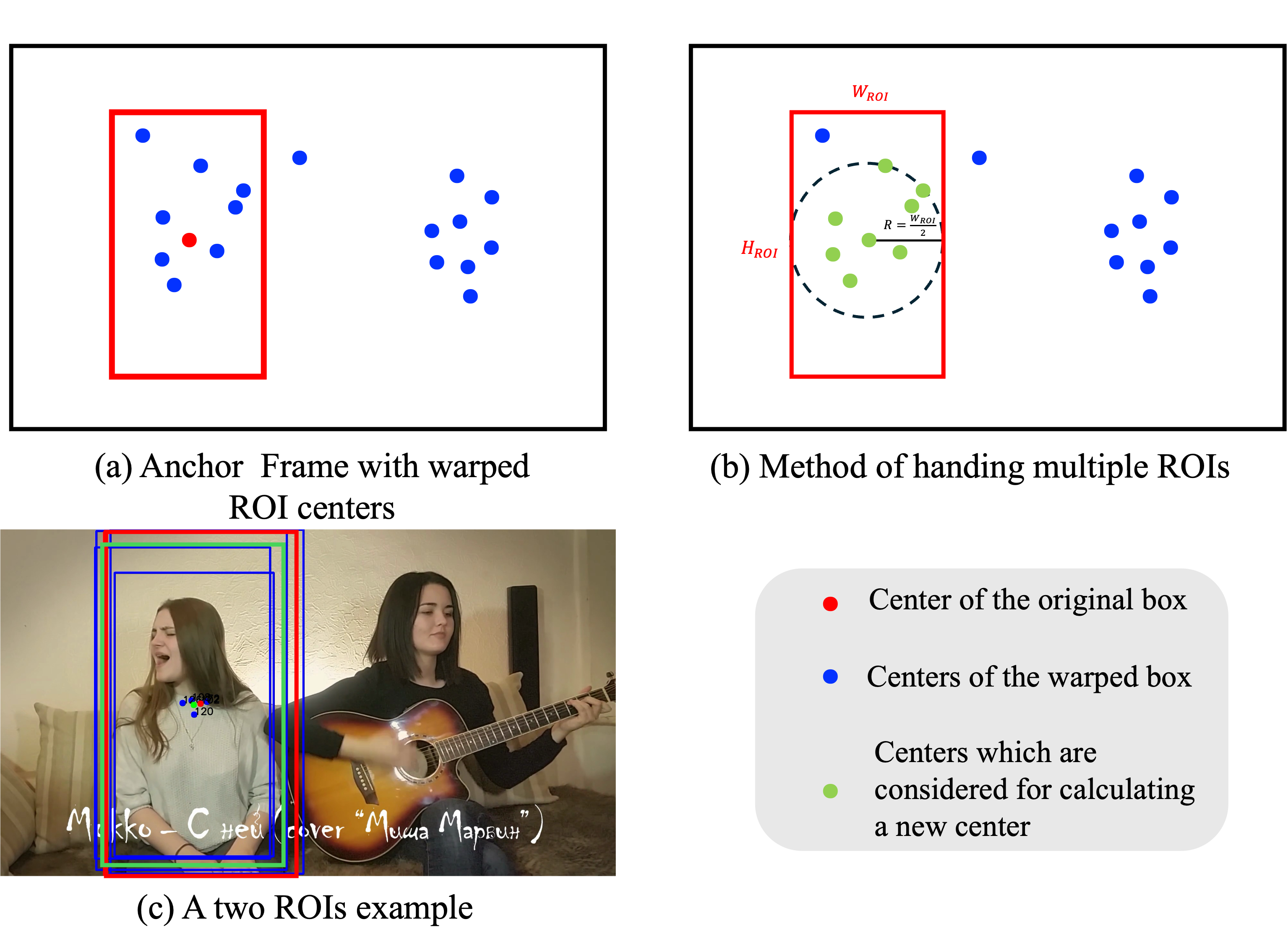}
\end{center}
\vspace{-10pt}
\caption{\footnotesize In (a), observe two clusters of ROI centers, with the original annotation located near one of the clusters. In (b), only use warped centers close to the original center. Specifically, define a radius equal to half the width of the original bounding box and use only points within this radius to compute a new weighted center. In (c), we illustrate a two-ROI example, demonstrating that the new weighted bounding box does not fall within the region of a different ROI.} 
\label{fig:long}
\label{fig:onecol}
\label{fig:mul_rois}
\vspace{-5pt}
\end{figure}

\noindent\textbf{Multiple ROIs Issue and Corner Case Handling.}
In real-world scenarios, the algorithm may encounter challenges due to some special cases. One such case involves occurrences of multiple regions of interest (ROIs), as illustrated in Fig. \ref{fig:mul_rois} (a). We have observed that ROI centers can form distinct clusters, arising from multiple similar and heavily overlapping ROIs. Another challenge arises in high-motion videos, where optical flow warping may produce inaccurate center predictions that deviate significantly from the true ROI. Such outliers must be filtered out.

%-----------------------------------------------------------%
%-----------------------------------------------------------%

To address these issues, we introduce the filtering strategy illustrated in Fig. \ref{fig:mul_rois} (b). Specifically, define a circular region centered at the anchor annotation's center, using half the annotation width as the radius. Only warped centers within this circle are considered by the weighting algorithm. As demonstrated in Fig. \ref{fig:mul_rois} (c), this method effectively resolves multi-ROI conflicts, ensuring that each final weighted bounding box remains within an appropriate ROI cluster.

Another common edge case is scene changes. To handle this, we leverage the PySceneDetect tool to precompute scene boundaries. During inter-frame temporal filtering, we only include warped centers from frames that fall within the same scene as the anchor annotation.

If a warped center meets any of the aforementioned conditions, we set its corresponding weight $w_{i} = 0$, effectively discarding its influence.

\subsection{Data Analysis}

Next, we analyze the statistics of the collected data in LIVE-YT VC, as well as those of LIVE-YT VC++, and compare the two.

\noindent\textbf{Inter-subject Labeling Consistency.}
We hypothesize that pairs of frames within a video should also have small distances between their box centers. A lower average distance between the box centers across a video indicates a higher degree of consistency among the subject labels. 

%-----------------------------------------------------------%
\begin{figure}[t]
\begin{center}
 \includegraphics[width=1.0\linewidth]{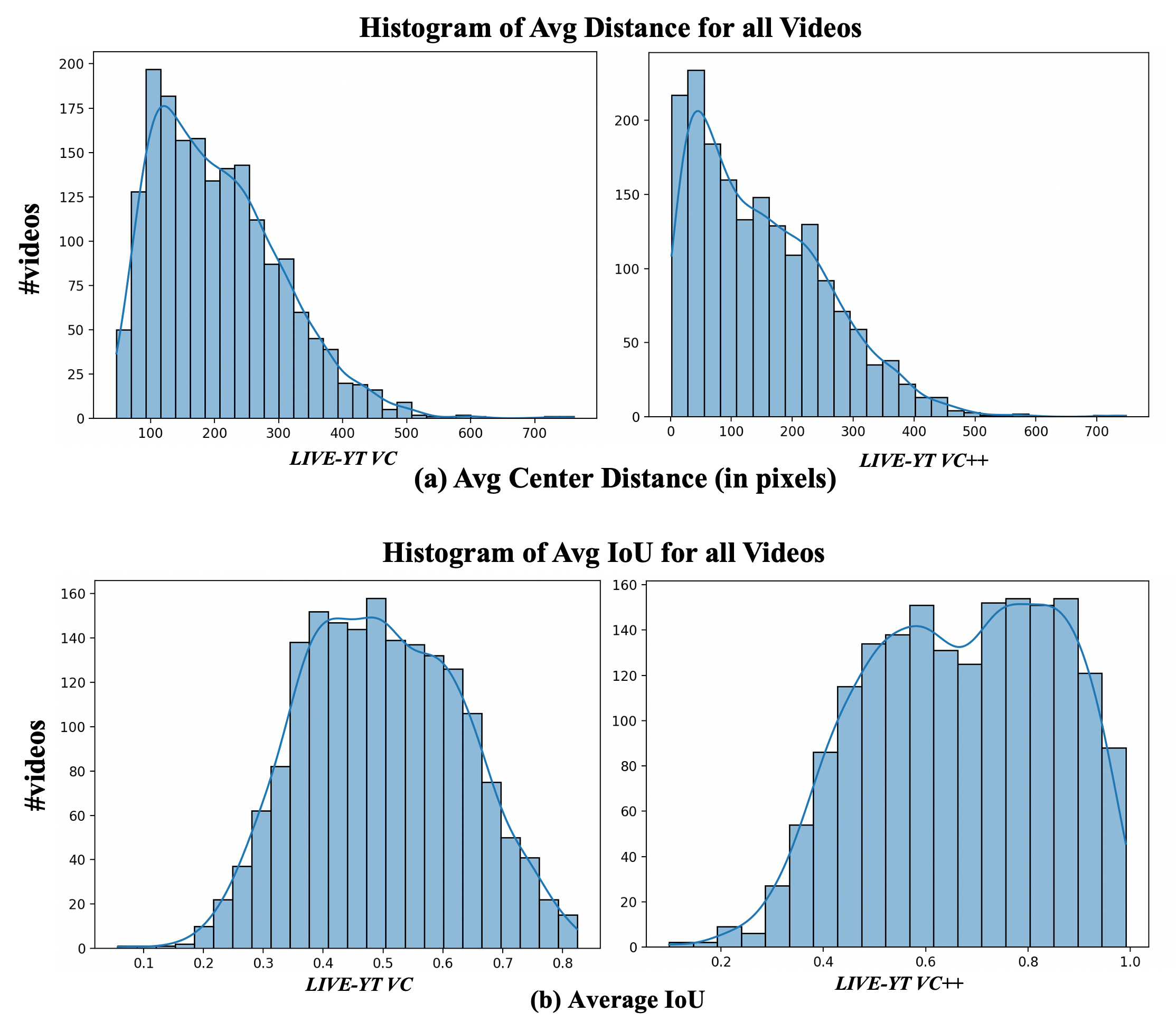}
\end{center}
\vspace{-10pt}
\caption{
%\footnotesize From (a) and (b), we observe that LIVE-YT VC++ tends to exhibit a skew toward higher average IoU values (left) and lower bounding box center distance values in pixels (right), indicating greater labeling consistency across the dataset compared to LIVE-YT VC. %From (c), we observe that LIVE-YT VC and LIVE-YT VC++ exhibit similar distributions of bounding box areas.
\footnotesize (a) shows the histograms of average center distances per video across the two datasets, while (b) plots the histograms of IoU overlaps between consecutive annotations. LIVE-YT VC++ exhibits improved performance against both metrics, indicating higher consistency.
}
\label{fig:new_stats}
\vspace{-5pt}
\end{figure}  
%-----------------------------------------------------------%
Fig. \ref{fig:new_stats} (a) shows the histogram of the box center differences over all videos in the LIVE-YT VC and LIVE-YT VC++ datasets. We noticed that the distribution is skewed toward lower values indicating a higher consistency of subject labels in LIVE-YT VC++. The average center distance in LIVE-YT VC is $205.1$ in pixels, but only $148.3$ in LIVE-YT VC++.

We would also expect the frames within a video to have high IoU, reflecting strong overlaps between the assigned bounding boxes. Since each frame in the original dataset was annotated by a different subject, we computed the IoU between consecutive frame pairs, then averaged the IoU values over each video. As shown in Fig. \ref{fig:new_stats} (b) (left diagram), most videos exhibit substantial overlap, with an overall average IoU of $0.50$. 

%-----------------------------------------------------------%
\begin{figure}[t]
\begin{center}
 \includegraphics[width=1.0\linewidth]{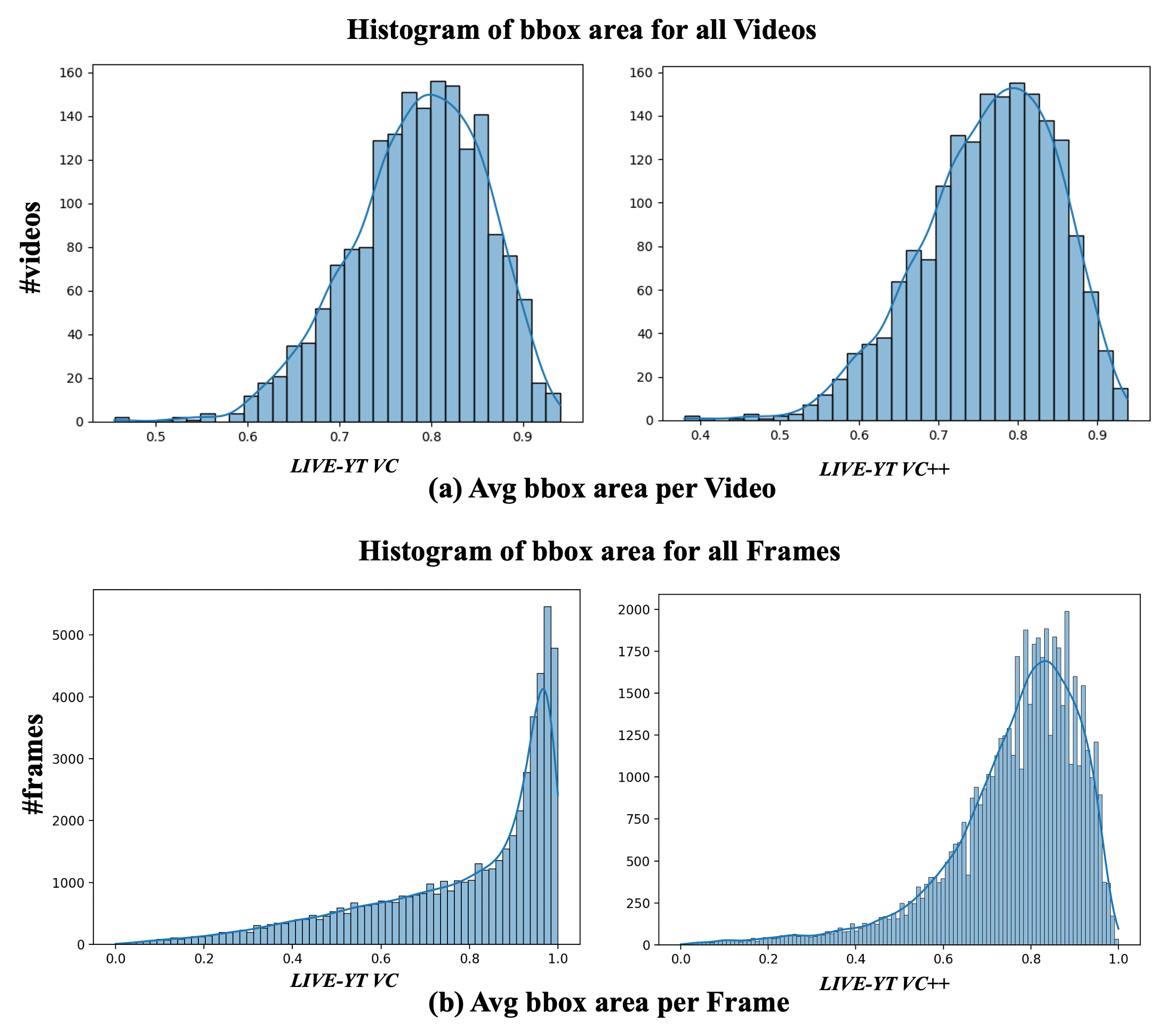}
\end{center}
\vspace{-10pt}
\caption{
%\footnotesize (a) presents the histogram of average bounding box sizes per video, where LIVE-YT VC and LIVE-YT VC++ show similar distributions. (b) displays the histogram of average bounding box sizes per frame, revealing that the distribution in LIVE-YT VC++ is more uniform compared to LIVE-YT VC.
\footnotesize We present the average bounding box area histograms per video (a) and per frame (b). The histogram for LIVE-YT VC++ is noticeably more uniform in the per-frame analysis, while the per-video results remain similar to the original dataset.
}
\label{fig:new_area}
\vspace{-5pt}
\end{figure}  
%-----------------------------------------------------------%

By contrast, as the processed videos in  LIVE-YT VC++ (Fig. \ref{fig:new_stats} (b) right), the distribution shifts noticeably toward higher IoU values: the average IoU across all videos increased to $0.67$, indicating improved temporal consistency.

\subsection{Box Area Analysis}

We also calculated the area of the bounding boxes in each frame and averaged the values across each video. From Fig. \ref{fig:new_area} (a), we observed that the per-video area distributions of the two datasets are quite similar. However, as shown in Fig. \ref{fig:new_area} (b), the per-frame analysis reveals that the box area histogram of LIVE-YT VC++ is more uniform than that of LIVE-YT-VC. This is because our algorithm aggregates adjacent annotations and helps reduce subject-specific bias in the labeling process.

\subsection{Visualization}

Next, we will show examples visualizing the new weighted box and warped boxes.

%-----------------------------------------------------------%
\begin{figure}[t]
\begin{center}
 \includegraphics[width=0.9\linewidth]{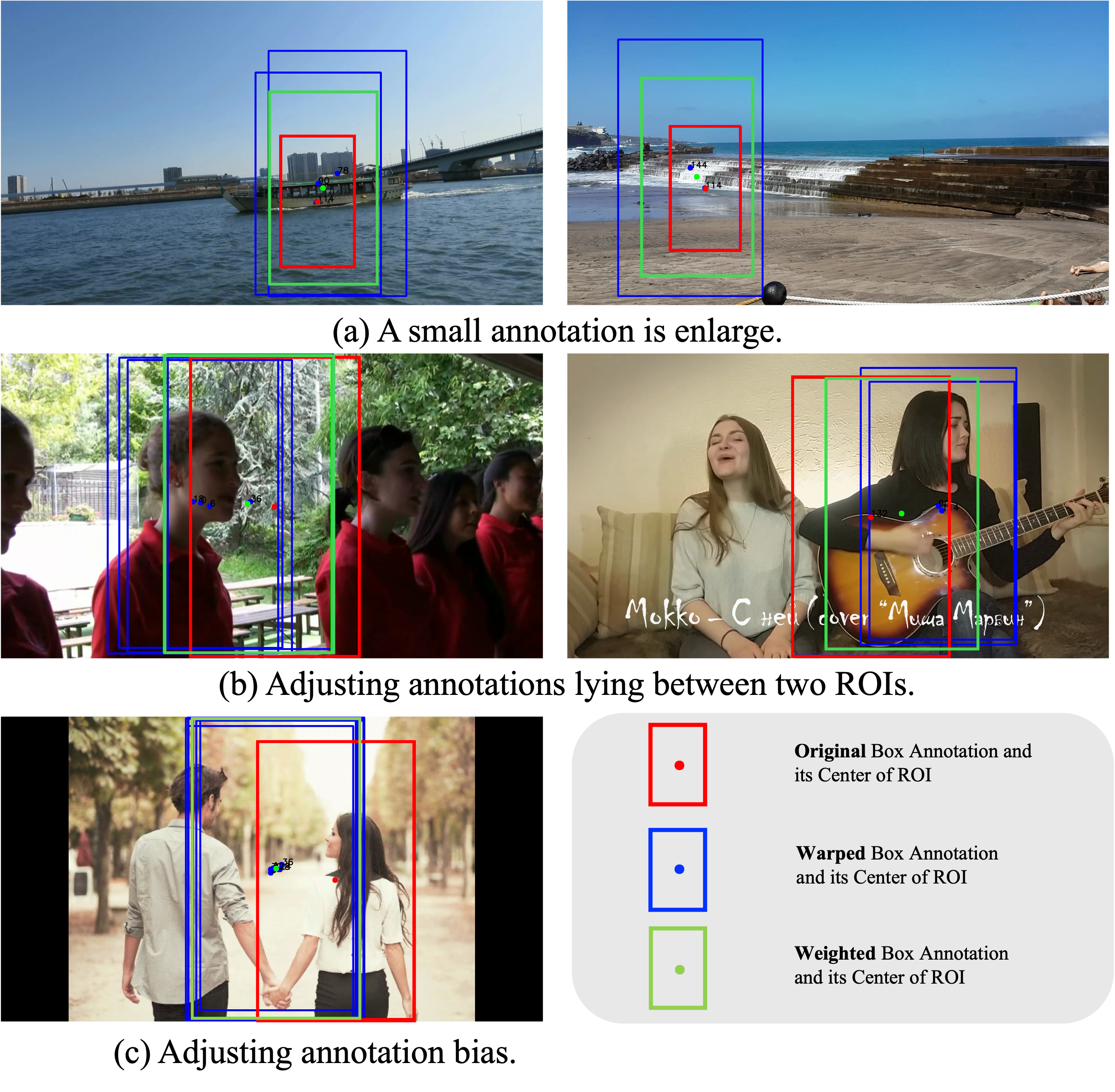}
\end{center}
\vspace{-10pt}
\caption{\footnotesize (a) illustrates how our algorithm can expand small annotations; (b) demonstrates its ability to adjust annotations that lie between two ROIs; and (c) shows how it can correct biased annotations.} 
\label{fig:++examples}
\vspace{-5pt}
\end{figure}
%-----------------------------------------------------------%
Fig. \ref{fig:++examples}, (a) shows that some annotations in LIVE-YT VC are quite small, but our algorithm effectively enlarges these to better capture the region of interest. In (b), one may observe annotations located between the two people (distinct ROIs); our method adjusts them to align more closely with a single ROI. In (c), most subjects labeled the couple's hands, while one subject (red) focused only on the girl. Our algorithm corrected this biased annotation by aligning it with the majority consensus.
\section{Experiments}\label{sec:experiments}
In this section, we present four experiments on both LIVE-YT VC and LIVE-YT VC++. First, we performed a saliency analysis to examine the relationship between visual saliency and our annotated bounding boxes. Second, we evaluated the utility of the new dataset by applying the saliency-based algorithm from SmartVidCrop \cite{apostolidis2021fast} and assessing its performance on our databases. Third, we repurpose two state-of-the-art video grounding models to test their accuracy and visual quality on the proposed datasets. Finally, we benchmark all methods—including naive cropping, SmartVidCrop, and the repurposed video grounding models—under the preserved-height setting.

\subsection{Implementation Details}
In the following subsections, we present several experiments. We first introduce the experimental setup.
The LIVE-YT-VC dataset contains 1,800 videos, which we divided it into 1,600 video pairs for training and 200 video pairs for testing.
For fair benchmarking, all non-training-free models were trained or fine-tuned over 7 epochs, with each experiment capped at approximately 2 hours. We used a batch size of 3, and all experiments were conducted on three NVIDIA A100 GPUs, each having 40GB of memory. The input frame resolution used in all experiments is 360p.

\subsection{Evaluation Protocols}
Following state-of-the-art video grounding methods, we employed m\_IoU and IoU@R as evaluation metrics. IoU calculates IoU over the set of frames, while m\_IoU is the average IoU across the test set. Further, IoU@R represents the percentage of samples in the test set with IoU exceeding a threshold R. For frame-wise aesthetic quality evaluation, we evaluate the perceived artistic and aesthetic value of each video frame using the LAION aesthetic predictor, trained on the LAION aesthetic subset \cite{schuhmann2022laion}. This predictor captures multiple aesthetic elements, such as layout, color richness and harmony, photorealism, naturalness, and the overall artistic quality of the frames. For temporal smoothness evaluation, we computed the mean absolute differences across frames. 

Average processing time refers to the average time taken to process a single video across the entire test set, calculated as the best result from three trials. These trials were conducted on machines free from other heavy-load processes. Our method, and the video grounding models were run on an NVIDIA A100 GPU (40 GB) server due to their high GPU memory requirements. SmartVidCrop \cite{apostolidis2021fast} was processed on an NVIDIA RTX 3090 GPU (24 GB) server.

To evaluate saliency detection, four common metrics were employed. LCC (Linear Correlation Coefficient) measures the linear relationship between predicted and ground-truth maps. SIM (Similarity) treats both maps as probability distributions and quantifies their overlap, reflecting the global distribution alignment. MAE (Mean Absolute Error) computes the average pixel-wise difference, providing a direct measure of prediction accuracy, while MSE (Mean Squared Error) measures pixel-wise differences but more greatly penalizes larger errors. Together, these metrics capture complementary aspects of saliency quality, from structural correlation to pixel-level fidelity.

\subsection{Relationship to Saliency Maps}
\begin{figure}[t]
\begin{center}
 \includegraphics[width=0.9\linewidth]{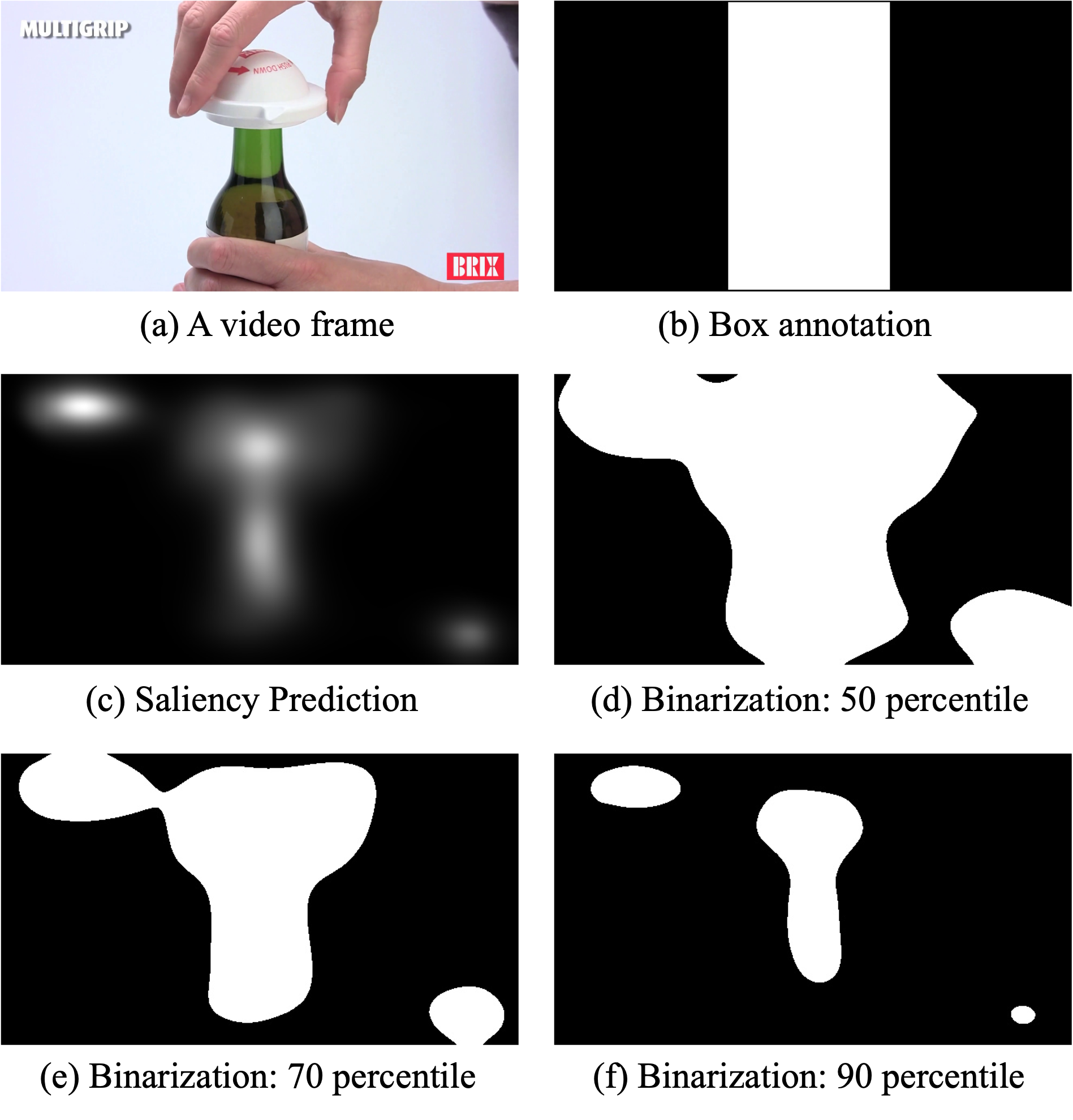}
\end{center}
\vspace{-10pt}
\caption{\footnotesize (a) An example video frame. (b) The bounding box annotation corresponding to (a) from LIVE-YT VC. (c) The saliency prediction corresponding to frame (a). (d) to (f) binarization of the saliency map in (c), using threshold values set at the 50th, 70th, and 90th percentiles, respectively.} 
\label{fig:long}
\label{fig:onecol}
\label{fig:sal}
\vspace{-5pt}
\end{figure}

 A saliency map is an image that highlights regions most likely to attract human gaze. To compare with our data, we employed UNISAL \cite{droste2020unified}, a state-of-the-art video saliency detection method, to extract saliency maps from sampled frames. We then evaluate their similarity to annotations in LIVE-YT VC and LIVE-YT VC++ using four metrics: LCC, SIM, MAE, and MSE. As shown in Table~\ref{sal}, both benchmarks exhibit moderate correlation with the predicted saliency maps (more so on LCC and SIM). Moreover, the enhanced LIVE-YT VC++ annotations achieved higher correlations than the original LIVE-YT VC, suggesting better alignment with visual attention. Fig.~\ref{fig:sal} illustrates an example of saliency map visualization on a single video frame. The original frame is shown in (a), with its corresponding bounding box annotation from LIVE-YT VC in (b). The predicted saliency map for frame (a) is shown in (c). Panels (d)–(f) depict the binarized versions of the saliency map in (c), obtained using threshold values at the 50th, 70th, and 90th percentiles, respectively.

%-----------------------------------------------------------%
\begin{table}
\caption{Correlation between saliency maps generated by UNISAL \cite{droste2020unified} on the LIVE-YT VC and LIVE-YT VC++ datasets.}
\label{sal}
\footnotesize
\begin{center}
\begin{tabular}{|c|c|c|c|c|} 
\hline
 Database & LCC (${\uparrow}$) &  SIM (${\uparrow}$) & MAE (${\downarrow}$) & MSE (${\downarrow}$) \\ 
\hline
\hline
LIVE-YT VC & 0.301 & 0.370 & 0.265 & 0.199 \\
\hline
LIVE-YT VC++ & 0.316 & 0.378 & 0.257 & 0.191 \\
\hline
\end{tabular}
\end{center}
\end{table}
%-----------------------------------------------------------%

%-----------------------------------------------------------%
%-----------------------------------------------------------%
\subsection{Relationship to Saliency Based Methods}
\begin{figure}[t]
\begin{center}
 \includegraphics[width=1.0\linewidth]{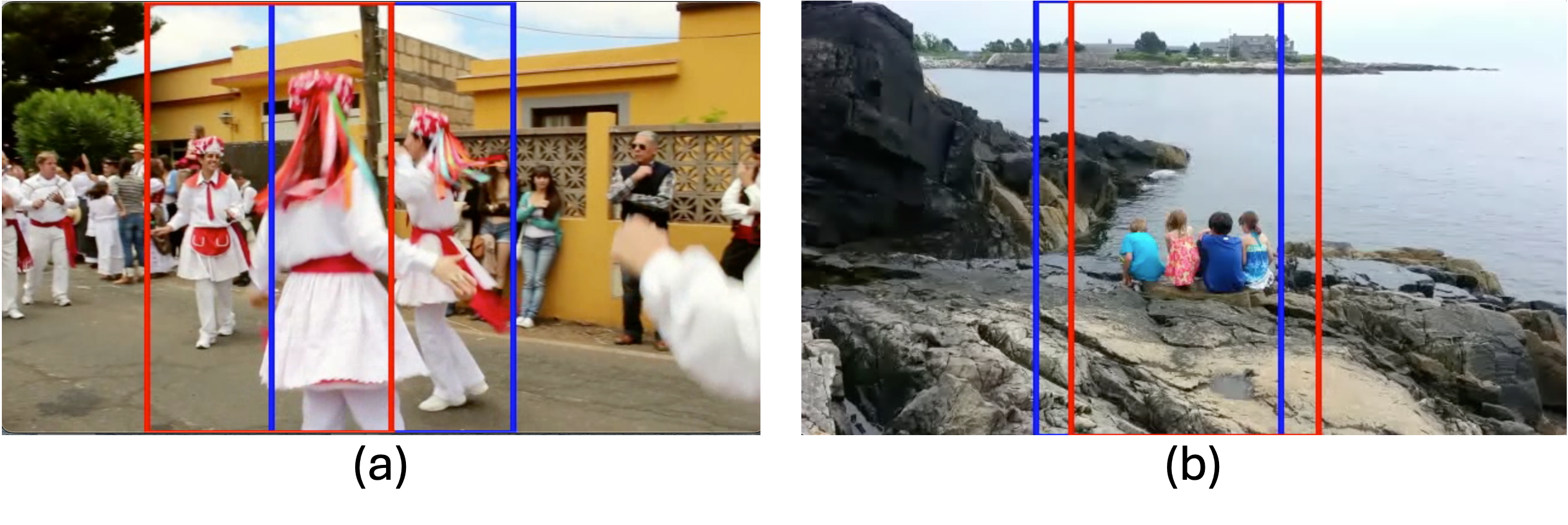}
\end{center}
\vspace{-10pt}
\caption{\footnotesize Comparison of bounding boxes generated by SmartVidCrop \cite{apostolidis2021fast} and human subjects. (a) Shows a low $vIoU \approx 0.30$ case.  (b) Shows a high $vIoU \approx 0.80$ case. Red boxes are ground truth labels from LIVE-YT VC and blue boxes are processing results of SmartVidCrop. Best viewed in color.}
\label{visual_box}
\vspace{-5pt}
\end{figure}

We leveraged a state-of-the-art video cropping model SmartVidCrop \cite{apostolidis2021fast} designed to use saliency maps. To ensure fair evaluation, all videos were resized to 360p, since SmartVidCrop was designed for this resolution. We also modified the bounding box labels of the videos in our database to scale to the video height while keeping the center intact, as SmartVidCrop lacks support for flexible height bounding box prediction. Table \ref{salvidcrop} presents the mean Intersection over Union (\textit{mIoU}) results. An \textit{mIoU} of $57.1$ on LIVE-YT VC and an \textit{mIoU} of $59.2$ on LIVE-YT VC++ indicates a significant degree of correlation between the bounding box labels in our database and the bounding boxes generated by SmartVidCrop. The correlations between the bounding boxes in LIVE-YT-VC++ and the results generated by SmartVidCrop were again higher. It is important to note that SmartVidCrop operates based on saliency maps, suggesting correlation between these maps and our bounding box labels. 
%The extended processing time is attributable to the multi-stage design of SmartVidCrop, which integrates deep learning models, machine learning algorithms, and video processing techniques. Furthermore, processing duration varies depending on factors such as video resolution and duration. 
%The relatively high \textit{mIoU} can be explained by the tendency of subjects to choose labels close to the central region as shown in Fig. \ref{fig:data_stats}. 
We analyzed the bounding boxes generated by SmartVidCrop to determine where it agrees or deviates from our subject labels. From our observations, videos containing multiple subjects, multiple scenes (collages), or simple scenes/backgrounds seem to generate lower \textit{IoU} predictions. We believe that on these kinds of videos, viewers are less likely to fixate to focus on a single region, hence lower agreements between ground truth and predicted labels. For visualization purposes, we selected frames from two videos and plotted two bounding boxes -- blue (predicted) and red (ground truth), as shown in Fig. \ref{visual_box}. We intentionally chose to highlight two cases, one with low and one with high agreement. Fig. \ref{visual_box}a shows an instance of a video containing multiple subjects. In this case the model selected a different crop than the subjects labeling the frame. By contrast, Fig. \ref{visual_box}b depicts a less complex video, having less ambiguity in the choiCe of RoI, resulting in high \textit{IoU}.

%-----------------------------------------------------------%
\begin{table}
\caption{Performance of SmartVidCrop \cite{apostolidis2021fast} on the LIVE-YT VC and LIVE-YT VC++ datasets.}
\label{salvidcrop}
\footnotesize
\begin{center}
\begin{tabular}{|c|c|} 
\hline
 Database & m\_IoU (${\uparrow}$) \\ 
\hline
\hline
LIVE-YT VC & 57.1 \\
\hline
LIVE-YT VC++ & 59.2 \\
\hline
\end{tabular}
\end{center}
\end{table}
%-----------------------------------------------------------%
%-----------------------------------------------------------%
%-----------------------------------------------------------%

\subsection{Benchmarking Video Grounding Models on LIVE-YT VC and LIVE-YT VC++}
\label{ssec:benchmark}
%As this is a newly emerging task in the AI era, there are currently no deep learning models specifically designed for it. To address this, we identified a closely related task—video grounding—which has several open-source models available. Video grounding aims to localize a target object within a video across both spatial and temporal dimensions. Due to the limited amount of training data available for our task, we leverage the knowledge embedded in pre-trained video grounding models and fine-tune them using our training pairs. Typically, video grounding models take two inputs: a landscape video and a text description that characterizes the target object. The output is a sequence of bounding boxes indicating the target’s location over time. However, our task does not require the use of a text backbone. Therefore, during the fine-tuning stage, we provide empty text inputs in all training pairs.
As this is a newly emerging task, there are currently no other competitive open-sourced deep learning models specifically designed for it. To address this, we identified a closely related task: video grounding, which has several open-source models available. Video grounding aims to localize a target object within a video across both spatial and temporal dimensions. Due to the limited amount of training data available for our task, we leverage the knowledge embedded in pre-trained video grounding models, and fine-tune them using our training pairs. Typically, video grounding models take two inputs: a landscape video and a text description that characterizes the target object. The output is a sequence of bounding boxes indicating the target's location over time. However, our task does not rely on text backbones or temporal grounding outputs. Therefore, during fine-tuning, we provide empty text inputs for all training samples. For temporal grounding, we set the start and end times to encompass the entire video duration in each data pair. Since annotations are only available on 30 sampled frames per video, we applied linear interpolation to generate dense labels across all frames.

\subsubsection{Introduction to  Video Grounding}
We introduce two representative video grounding datasets along with prior state-of-the-art models designed for this task.

\noindent\textbf{Datasets.}
HCSTVG \cite{tang2021human}, which focuses exclusively on humans in videos, is available in two versions: HCSTVG-v1 and an expanded version:  HCSTVG-v2. The videos in these datasets were collected from cinematic movies. The durations are normalized to 20 seconds. All the text descriptions in the dataset are declarative. VidSTG \cite{zhang2020does} is another database built from a video relation dataset, containing 99,943 sentence descriptions of 79 types of queried objects (including humans, animals, and everyday objects). The average video duration is 28.01 seconds, with an average temporal length of tracked object tubes of 9.68 seconds. The average lengths of the declarative and interrogative sentences are 11.12 and 8.98 seconds, respectively. 

\noindent\textbf{Related Works.}
Video grounding aims to generate a spatio-temporal tube based on a given text query. Early two-stage methods such as \cite{zhang2020does, zhang2020object}, leverage pre-defined detectors. Recently, emerging one-stage designs, including \cite{yang2022tubedetr, jin2022embracing, lin2023collaborative, gu2024context}, have enabled end-to-end processing of videos. In the following experiments, we adapt two recent state-of-the-art models, STCAT \cite{jin2022embracing} and CG-STVG \cite{gu2024context}, to our video cropping task.

%-----------------------------------------------------------%
\begin{table*}
\begin{center}
\caption{Quantitative results of repurposed state-of-the-art video grounding models on the LIVE-YT-VC database.}
\label{liveytvc}
\begin{tabular}{| c | c |  c c c | c c | c |} 
\hline
 Method & Pre-Training Data & m\_IoU (${\uparrow}$) & IoU@0.3 (${\uparrow}$) & IoU@0.5 (${\uparrow}$) & Aesthetic (\%${\uparrow}$) & Temporal (\%${\uparrow}$) & Avg. Time (sec${\downarrow}$) \\ 
\hline
\hline
 & N/A & 45.0 & 86.5 & 37.5 & 46.1 & \textbf{95.8} &  \\
STCAT \cite{jin2022embracing} & HCSTVG-v1 \cite{tang2021human} & 52.3 & 96.0 & 55.0 & 46.4 & 93.8 & \textbf{3.4} \\
 & VidSTG \cite{zhang2020does} & 49.5 & 93.0 & 45.5 & 46.1 & 93.6 &  \\
\hline
 & N/A & 46.6 & 89.0 & 39.5 & 46.4 & \textbf{95.8} & \\
CG-STVG \cite{gu2024context} & HCSTVG-v2 \cite{tang2021human} & \textbf{53.1} & \textbf{96.0} & \textbf{59.5} & 46.6 & 93.7  & 4.7 \\
 & VidSTG \cite{zhang2020does} & 52.4 & 95.0 & 55.5 & \textbf{46.7} & 93.7  & \\
\hline
GT & N/A & - & - & - & 46.7 & 90.4 & - \\
\hline
\end{tabular}
\end{center}
\end{table*}
%-----------------------------------------------------------%
%-----------------------------------------------------------%
\begin{table*}
\begin{center}
\caption{Quantitative results of repurposed state-of-the-art video grounding models on the LIVE-YT-VC++ database.}
\label{liveytvc2}
\begin{tabular}{| c | c | c c c | c c | c |} 
\hline
 Method & Pre-Training Data & m\_IoU (${\uparrow}$) & IoU@0.3 (${\uparrow}$) & IoU@0.5 (${\uparrow}$) & Aesthetic (\%${\uparrow}$) & Temporal (\%${\uparrow}$) & Avg. Time (sec${\downarrow}$) \\ 
\hline
\hline
 & N/A & 45.0 & 80.5 & 41.5 & 45.1 & \textbf{95.8} &  \\
STCAT \cite{jin2022embracing} & HCSTVG-v1 \cite{tang2021human} & 58.2 & 95.0 & 67.0 & 46.4 & 94.0 & \textbf{3.4} \\
 & VidSTG \cite{zhang2020does} & 54.8 & 94.0 & 64.0 & 46.0 & 93.7 &  \\
\hline
 & N/A & 49.4 & 88.5 & 48.0 & 46.0 & \textbf{95.8} &  \\
CG-STVG \cite{gu2024context} & HCSTVG-v2 \cite{tang2021human} & \textbf{59.1} & 96.0 & \textbf{72.0} & 46.5 & 93.7  & 4.7 \\
 & VidSTG \cite{zhang2020does} & 58.8 & \textbf{96.5} & 69.0 & \textbf{46.6} & 93.8  &  \\
\hline
GT & N/A & - & - & - & 46.8 & 92.8 & - \\
\hline
\end{tabular}
\end{center}
\end{table*}
%-----------------------------------------------------------%

%-----------------------------------------------------------%
\begin{table*}
\begin{center}
\caption{Quantitative results of repurposed state-of-the-art video grounding models on the LIVE-YT-VC database, with the height of the cropping regions preserved. (Our implementation of SmartVidCrop \cite{apostolidis2021fast} follows the original setting and evaluated on 360p)}
\label{liveytvcpreserve}
\begin{tabular}{| c | c | c c c | c c | c |} 
\hline
 Method & Pre-Training Data & m\_IoU (${\uparrow}$) & IoU@0.3 (${\uparrow}$) & IoU@0.5 (${\uparrow}$) & Aesthetic (\%${\uparrow}$) & Temporal (\%${\uparrow}$) & Avg. Time (sec${\downarrow}$) \\ 
\hline
\hline
Center Crop & Training-Free & 55.7 & 93.0 & 65.0 & 46.7 & \textbf{95.8} & \textbf{2.3} \\
SmartVidCrop \cite{apostolidis2021fast} & Training-Free & 57.1 & 93.5 & 63.5 & 42.6 & 94.9 & 8.9 \\
STCAT \cite{jin2022embracing} & HCSTVG-v1 \cite{tang2021human} & \textbf{65.4} & \textbf{99.5} & \textbf{87.0} & 47.5 & 94.3 & 3.4 \\
CG-STVG \cite{gu2024context} & HCSTVG-v2 \cite{tang2021human} & \textbf{65.4} & 99.0 & 85.0 & \textbf{47.6} & 94.2  & 4.7 \\
\hline
GT & N/A & - & - & - & 47.6 & 91.8 & - \\
\hline
\end{tabular}
\end{center}
\end{table*}
%-----------------------------------------------------------%

%-----------------------------------------------------------%
\begin{table*}
\begin{center}
\caption{Quantitative results of repurposed state-of-the-art video grounding models on the LIVE-YT-VC++ database, with the height of the cropping regions preserved. (Our implementation of SmartVidCrop \cite{apostolidis2021fast} follows the original setting and evaluated on 360p)}
\label{liveytvcpreserve2}
\begin{tabular}{| c | c | c c c | c c | c |} 
\hline
 Method & Pre-Training Data & m\_IoU (${\uparrow}$) & IoU@0.3 (${\uparrow}$) & IoU@0.5 (${\uparrow}$) & Aesthetic (\%${\uparrow}$) & Temporal (\%${\uparrow}$) & Avg. Time (sec${\downarrow}$) \\ 
\hline
\hline
Center Crop & Training-Free & 57.5 & 92.5 & 66.0 & 46.7 & \textbf{95.8} & \textbf{2.3} \\
SmartVidCrop \cite{apostolidis2021fast} & Training-Free & 59.2 & 93.5 & 68.5 & 42.6 & 94.9 & 8.9 \\
STCAT \cite{jin2022embracing} & HCSTVG-v1 \cite{tang2021human} & 68.8 & 99.0 & 87.5 & \textbf{47.6} & 94.5 & 3.4 \\
CG-STVG \cite{gu2024context} & HCSTVG-v2 \cite{tang2021human} & \textbf{69.5} & \textbf{99.5} & \textbf{89.5} & \textbf{47.6} & 94.2 & 4.7 \\
\hline
GT & N/A & - & - & - & 47.7 & 93.4 & - \\
\hline
\end{tabular}
\end{center}
\end{table*}
%-----------------------------------------------------------%
\subsubsection{Benchmarking under Non-height-preserving Scenarios}
Table \ref{liveytvc} and Table \ref{liveytvc2} present benchmarking results under the non-preserved height scenario on both LIVE-YT VC and LIVE-YT VC++. The outcomes indicate that pre-trained weights play a crucial role in enabling model convergence. Models trained from scratch generally fail to converge within a limited number of epochs, resulting in outputs biased toward the center of the frame and poor temporal smoothness. By contrast, using HCSTVG-v2 as pre-trained weights consistently leads to better accuracy. Among the video-language models evaluated, CG-STVG \cite{gu2024context}, which employs a stronger video encoder and decoder, tends to achieve higher accuracy than STCAT \cite{jin2022embracing}, but lower efficiency due to its complex design.
In the final row of the ground truth perceptual quality results, LIVE-YT VC++ may be observed to yield superior performance compared to LIVE-YT VC.

\subsubsection{Benchmarking under Height-preserving Scenarios}
In height-preserving scenarios, the cropped region is constrained to the $x$-axis, while the $y$-axis remains unaffected. Table \ref{liveytvcpreserve} and Table \ref{liveytvcpreserve2} present benchmarking results for the height-preserving scenario on both LIVE-YT VC and LIVE-YT VC++. We introduce a common baseline, the eponymous Center Crop, which achieves the best temporal smoothness and the lowest processing time. The Center Crop baseline also yielded comparable \textit{mIoU} to SmartVidCrop, which may be attributed to the tendency of annotators to select regions near the image center, as illustrated in Fig.~\ref{fig:data_stats}. Another multi-stage baseline is SmartVidCrop \cite{apostolidis2021fast}, which showed a slight improvement in \textit{mIoU} over Center Crop. However, its extended processing time arises from its multi-stage design, integrating deep learning models, machine learning algorithms, and video processing techniques. In addition, the processing duration varies with factors such as video resolution and length.

Compared to the repurposed STCAT \cite{jin2022embracing} and CS-STVG \cite{gu2024context} models, both Center Crop and SmartVidCrop significantly underperformed in terms of box accuracy. Among the evaluated video-language models, CG-STVG and STCAT  exhibit similar accuracy, likely because this task is simpler than the previous one. In the final row of the ground truth perceptual quality results, LIVE-YT VC++ may be observed to have delivered superior performance compared to LIVE-YT VC.
\section{Conclusion}
\label{sec:refs}
%This paper contributes to three key areas. Firstly, we conducted an extensive subjective study on video portrait region cropping, involving 90 subjects. Secondly, we introduced the largest database for subjective video portrait region cropping, comprising 1800 videos. Additionally, a thorough data analysis was conducted, including outlier analysis, labeling consistency analysis, and bounding box statistic analysis. Lastly, we evaluated the effectiveness of the SmartVidCrop algorithm on the new database to assess its performance. These contributions collectively advance understanding and methodologies in the domain of video aspect ratio transformation and cropping. Looking ahead, we aim to develop a novel end-to-end deep learning-based model using this database to further improve solutions to this problem.
We have made several contributions to the field of video cropping and subjective annotation. First, we conducted the largest subjective video portrait region cropping study, involving 90 subjects and 54,000 ratings. Based on the results, we created the largest subjective video portrait region cropping database, LIVE-YT VC, consisting of 1,800 videos and 324,000 annotated frames. To further enhance the dataset, we modified the annotations to create LIVE-YT VC++, by applying an novel intra-frame temporal filter to smooth annotations. We demonstrated the usefulness of these new perceptual data resources by evaluating the SmartVidCrop algorithm on them. We also performed a saliency analysis to investigate possible relationships between visual saliency and our annotations. Finally, we repurposed state-of-the-art video grounding models to establish an end-to-end video cropping framework, achieving strong performance across accuracy and visual quality metrics. It is our hope that these contributions will significantly advance the understanding and capabilities of video cropping in real-world applications.

\bibliography{reference}
\bibliographystyle{IEEEtran}

\end{document}